\documentclass[sigconf]{acmart}

\AtBeginDocument{%
  }

\copyrightyear{2026}
\acmYear{2026}
\setcopyright{cc}
\setcctype{by}
\acmConference[MM '26]{Proceedings of the 34th ACM International Conference on Multimedia}{November 10--14, 2026}{Rio de Janeiro, Brazil}
\acmBooktitle{Proceedings of the 34th ACM International Conference on Multimedia (MM '26), November 10--14, 2026, Rio de Janeiro, Brazil}
\acmDOI{10.1145/3767308.3836069}
\acmISBN{979-8-4007-2213-4/2026/11}
\usepackage{xspace}
\usepackage{graphicx}
\usepackage{multirow}
\usepackage{colortbl}
\usepackage{subcaption}
\definecolor{MyRed}{HTML}{D2042D}
\definecolor{MyGreen}{HTML}{00A36C}
\definecolor{MyBlue}{HTML}{1F51FF}
\definecolor{MyPurple}{HTML}{AF58BA}
\definecolor{MyYellow}{HTML}{FFC61E}
\definecolor{MyOrange}{HTML}{F28522}
\definecolor{MyGrey}{HTML}{A0B1BA}
\definecolor{MyBrown}{HTML}{A6761D}
\definecolor{lightgrey}{rgb}{0.83, 0.83, 0.83}
\definecolor{lavendermist}{rgb}{0.9, 0.9, 0.98}
\definecolor{lightblue}{HTML}{dfebf7}
\definecolor{lightgreen}{HTML}{f8fcf4}

\newcommand{\grey}[1]{{\color{MyGrey}#1}}

\newcommand{\na}{\grey{N/A}}

\newcommand{\model}{\textsc{\texttt{MMaDA-VLA}}\xspace}
\newcommand{\mask}{\textsc{\texttt{[M]}}\xspace}
\newcommand{\tgx}[1]{\grey{#1}}

\begin{document}

\title[MMaDA-VLA]{MMaDA-VLA: Large Diffusion Vision-Language-Action Model with Unified Multi-Modal Instruction and Generation}

\author{Yang Liu}
\email{liuyang67@westlake.edu.cn}
\affiliation{%
  \institution{Westlake University}
  \city{Hangzhou}
  \country{China}}

\author{Pengxiang Ding}
\email{dingpx2015@gmail.com}
\affiliation{%
  \institution{Westlake University}
  \institution{Zhejiang University}
  \city{Hangzhou}
  \country{China}}

\author{Tengyue Jiang}
\email{23013246@mail.ecust.edu.cn}
\affiliation{%
  \institution{East China University of Science and Technology}
  \city{Shanghai}
  \country{China}}

\author{Xudong Wang}
\email{wangxudong32@huawei.com}
\affiliation{%
  \institution{Huawei Technologies Ltd.}
  \city{Hangzhou}
  \country{China}}

\author{Wenxuan Song}
\email{songwenxuan0115@gmail.com}
\affiliation{%
  \institution{The Hong Kong University of Science and Technology (Guangzhou)}
  \city{Guangzhou}
  \country{China}}

\author{Minghui Lin}
\email{linminghui1226@gmail.com}
\affiliation{%
  \institution{Westlake University}
  \city{Hangzhou}
  \country{China}}

\author{Han Zhao}
\email{zhaohan34@westlake.edu.cn}
\affiliation{%
  \institution{Westlake University}
  \institution{Zhejiang University}
  \city{Hangzhou}
  \country{China}}

\author{Hongyin Zhang}
\email{zhanghongyin@westlake.edu.cn}
\affiliation{%
  \institution{Westlake University}
  \institution{Zhejiang University}
  \city{Hangzhou}
  \country{China}}

\author{Zifeng Zhuang}
\email{zhuangzifeng@westlake.edu.cn}
\affiliation{%
  \institution{Westlake University}
  \city{Hangzhou}
  \country{China}}

\author{Wei Zhao}
\email{zhaowei@westlake.edu.cn}
\affiliation{%
  \institution{Westlake University}
  \city{Hangzhou}
  \country{China}}

\author{Siteng Huang}
\email{siteng.huang@gmail.com}
\affiliation{%
  \institution{Zhejiang University}
  \city{Hangzhou}
  \country{China}}

\author{Jinkui Shi}
\email{shijinkui@huawei.com}
\affiliation{%
  \institution{Huawei Technologies Ltd.}
  \city{Hangzhou}
  \country{China}}

\author{Donglin Wang}
\email{wangdonglin@westlake.edu.cn}
\correspondingauthor
\affiliation{%
  \institution{Westlake University}
  \city{Hangzhou}
  \country{China}}

\renewcommand{\shortauthors}{Liu et al.}

\begin{abstract}

Vision-Language-Action (VLA) models map visual observations and natural-language instructions to robot actions; however, hierarchical and autoregressive paradigms often incur architectural overhead, accumulate long-horizon errors, and require auxiliary modules to capture environment dynamics.
To this end, we present \model, a fully native, pretrained discrete diffusion VLA that unifies multi-modal understanding and generation.
Specifically, \model uses a shared discrete token space to jointly denoise a future goal observation and an action chunk, grounding actions in predicted visual outcomes without an auxiliary world model.
In this way, parallel, order-free refinement improves long-horizon consistency.
Extensive experiments and comprehensive analyses demonstrate that \model achieves an average success rate of 98.0\% on LIBERO and an average successful sequence length of 4.78 on CALVIN, while performing strongly in real-world settings.
The project page is available at \url{https://yliu-cs.github.io/MMaDA-VLA}.

\end{abstract}



\maketitle

\begin{figure*}[t]
  \includegraphics[width=\textwidth]{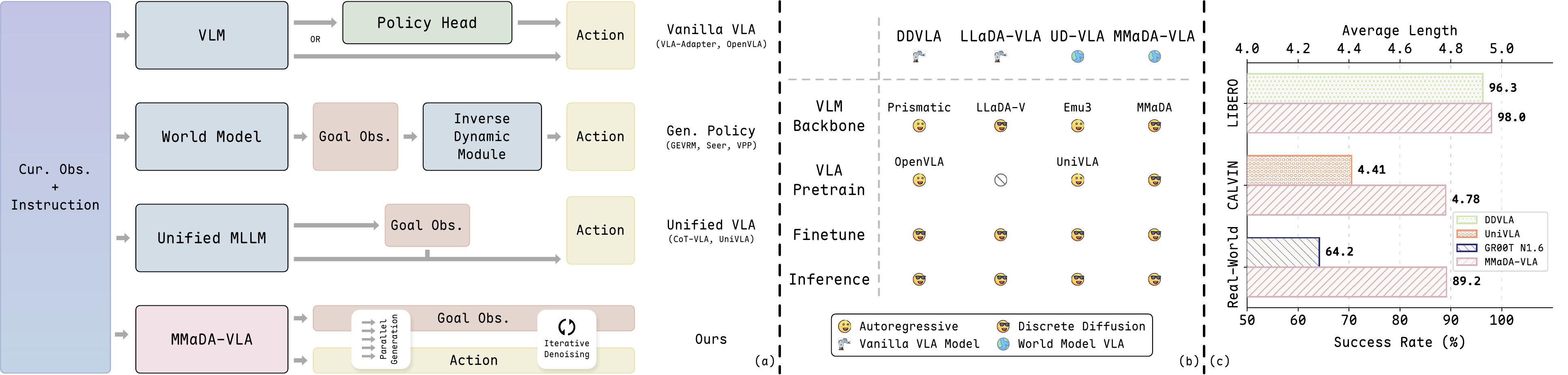}
  \caption{(a) Comparison of conventional VLA methods and \model. (b) Comparison with discrete diffusion methods. (c) Performance comparison with state-of-the-art methods. Cur., Obs., Gen., and DDVLA denote current, observation, generation, and Discrete Diffusion VLA, respectively.}
  \Description{Three panels compare vision-language-action approaches. The left panel contrasts policy-head, world-model, unified-VLA, and MMaDA-VLA pipelines. The middle panel shows which competing models use autoregressive or discrete-diffusion backbones, VLA pre-training, fine-tuning, and inference. The right panel reports MMaDA-VLA at 98.0 percent on LIBERO, 4.78 average length on CALVIN, and 89.2 percent on real-world tasks.}
  \label{fig:teaser}
\end{figure*}

\section{Introduction}
\label{sec:intro}

Vision-Language-Action (VLA) models aim to control robots to perform manipulation tasks based on visual observations and natural-language instructions.
Robotic manipulation is central to many applications, including household assistance, industrial automation, and rescue support, where both generalization and precision are required.
Most state-of-the-art (SOTA) VLA models \citep{DBLP:conf/rss/BrohanBCCDFGHHH23, DBLP:conf/corl/KimPKXB0RFSVKBT24, DBLP:journals/corr/abs-2501-09747, DBLP:journals/corr/abs-2502-19645, DBLP:journals/corr/abs-2509-09372} follow either a hierarchical paradigm that augments existing vision-language models (VLMs) with a dedicated policy head or an end-to-end approach that discretizes actions and generates them autoregressively.
While these methods benefit from the strong representation capabilities of modern VLMs and often demonstrate effective action prediction and environment generalization, hierarchical designs introduce additional architectural complexity and training costs and can reduce information fidelity across module boundaries.
In contrast, action-discretization approaches may generate action sequences with weak temporal consistency, limiting global trajectory planning and leading to compounding errors over long horizons.
Moreover, both paradigms generally lack an explicit mechanism for modeling environment dynamics through future-observation prediction.

These issues are particularly challenging for several reasons, including the reliance on continuous action policy heads to meet the precision requirements of manipulation and the dependence on existing VLMs to achieve generalist behavior.
To address these challenges, prior work has explored several directions.
First, action quantization schemes discretize continuous actions while preserving reconstruction accuracy \citep{DBLP:journals/corr/abs-2501-09747, Wang_2025_ICCV, DBLP:journals/corr/abs-2510-09667}.
Second, specialized decoding strategies enable multi-step action prediction in a single forward pass and parallel decoding for otherwise autoregressive models \citep{DBLP:journals/corr/abs-2502-19645, DBLP:conf/iros/SongCDZZZGLWWML25, DBLP:journals/corr/abs-2506-13725}.
Third, additional visual-generation modules or auxiliary objectives model environment dynamics by predicting future observations or learning inverse dynamics \citep{DBLP:conf/iclr/TianYZWL0P25, DBLP:conf/icml/abs-2501-18867, DBLP:journals/corr/abs-2506-19850, DBLP:journals/corr/abs-2506-21539, DBLP:journals/corr/abs-2507-04447}.
Despite these advances, as shown in Figure~\ref{fig:teaser} (a), many approaches still incur substantial design and computational overhead, require more complex architectures, and propagate errors across multi-stage generation.
More recently, as depicted in Figure~\ref{fig:teaser} (b), unified VLA models based on large diffusion backbones have replaced autoregressive decoding with masked-token prediction \citep{DBLP:journals/corr/abs-2508-20072, DBLP:journals/corr/abs-2509-06932} and incorporated goal-image generation before action prediction to better capture environment dynamics \citep{DBLP:journals/corr/abs-2511-01718}.
However, these methods are typically fine-tuned from autoregressive models. This adaptation creates inconsistencies between training and inference, thereby limiting performance.
To address this limitation, we propose \model, a fully native, pretrained large diffusion VLA model that unifies multi-modal understanding and generation in a single framework.

Specifically, \model learns an end-to-end, temporally consistent, and dynamics-aware policy without auxiliary world models or training-inference mismatches.
Our main contribution is a native discrete diffusion VLA formulation that represents language, images, and continuous robot controls within a single discrete token space and trains a single backbone with a masked-token denoising objective to jointly generate a future goal observation and an action chunk in parallel.
This design is effective because iterative denoising supports global, order-free refinement, avoiding an arbitrary autoregressive ordering over inherently unordered action dimensions and allowing the model to repeatedly align the action chunk with the predicted future visual outcome. This process encourages an implicit representation of task-relevant state evolution.
Compared with hierarchical VLA designs and autoregressive action decoding, \model provides a simpler architecture and training procedure, reduces error accumulation through parallel refinement, improves long-horizon consistency through chunked action prediction, and strengthens dynamics grounding through goal observation generation, all without relying on external generative modules.
We evaluate \model against a range of baselines on the LIBERO \citep{DBLP:conf/nips/LiuZGFLZS23} and CALVIN \citep{DBLP:journals/ral/MeesHRB22} simulation benchmarks and in real-world tasks.
As shown in Figure~\ref{fig:teaser} (c), \model achieves state-of-the-art performance in these settings, with a 98.0\% average success rate on LIBERO and an average completed sequence length of 4.78 on CALVIN under the ABC$\rightarrow$D setting. Extensive experiments and analyses further demonstrate the effectiveness and broad applicability of the proposed method.

Overall, our main contributions are listed as follows: \\
$\bullet$ We introduce \model, a fully native, pretrained large diffusion VLA model that unifies multi-modal instruction and generation. \\
$\bullet$ We develop an end-to-end training pipeline combining large-scale cross-embodiment pre-training with task-specific fine-tuning. \\
$\bullet$ Extensive simulation and real-world experiments demonstrate that \model consistently outperforms prior SOTA VLAs.

\section{Related Work}
\label{sec:rel}

\subsection{Large Diffusion Models}

By generating text token by token, large language models (LLMs) \citep{Radford2018ImprovingLU, Radford2019LanguageMA, conf/nips/BrownMRSKDNSSAA20, journals/corr/abs-2302-13971, DBLP:journals/corr/abs-2309-16609, DBLP:journals/corr/abs-2501-12948} achieve strong performance across a broad range of tasks.
Nonetheless, several limitations can be traced directly to the left-to-right factorization underlying autoregressive generation.
In particular, this sequential constraint can hinder bidirectional reasoning and cause systematic generalization failures on reversal tasks \citep{DBLP:conf/iclr/BerglundTKBSKE24}.
In parallel, diffusion models have shown excellent scalability and sample fidelity in continuous domains such as images and audio \citep{DBLP:conf/iccv/ZhangRA23, DBLP:conf/iclr/BenitaEK24, DBLP:conf/iclr/YuLGVSMCGGHG0ER24, DBLP:conf/nips/DhariwalN21, DBLP:conf/nips/HoJA20, DBLP:conf/nips/LiHRMM23}, motivating efforts to adapt diffusion-style generation to discrete language tokens.
Recent work develops discrete diffusion language models by defining a forward noising process, often implemented with token masking, and learning a reverse denoising process to reconstruct clean text \citep{nie2025large, DBLP:journals/corr/abs-2508-15487, DBLP:journals/corr/abs-2505-19223, DBLP:journals/corr/abs-2510-04146, DBLP:journals/corr/abs-2509-24389, DBLP:journals/corr/abs-2510-06303}.
These diffusion models support parallel prediction of multiple tokens across denoising steps, providing a promising alternative for next-generation LLMs.
Beyond language, diffusion models have also demonstrated strong potential in multi-modal modeling \citep{DBLP:journals/corr/abs-2505-16839, DBLP:journals/corr/abs-2505-16933, DBLP:journals/corr/abs-2505-16990, yang2025mmada}, where their non-strictly sequential modeling aligns with the demands of vision.

\subsection{Vision-Language-Action Models}

Vision-Language-Action (VLA) models built on vision-language models (VLMs) provide a new approach to controlling robots across diverse everyday tasks.
Existing VLA models can be broadly categorized into two main families: discrete VLA models \citep{DBLP:conf/rss/BrohanBCCDFGHHH23, DBLP:conf/corl/ZitkovichYXXXXW23, DBLP:conf/corl/KimPKXB0RFSVKBT24, DBLP:journals/corr/abs-2501-09747,
DBLP:conf/iros/SongCDZZZGLWWML25} and hierarchical continuous VLA models \citep{DBLP:journals/corr/abs-2502-19645, DBLP:journals/corr/abs-2410-24164, DBLP:journals/corr/abs-2412-14058, DBLP:journals/corr/abs-2509-09372}.
Discrete VLA models typically generate action tokens by extending the original VLM vocabulary during fine-tuning, thereby casting action prediction as token generation.
In contrast, hierarchical continuous VLA models generally augment the VLM with a policy head or action expert module that directly predicts continuous actions.
Moreover, these modules can be decoupled and executed asynchronously to increase the control frequency \citep{DBLP:journals/corr/abs-2410-08001, DBLP:journals/corr/abs-2505-03912}.

\subsection{World Models for Robotics}

Recently, advances in visual generation have driven rapid progress in world models \citep{DBLP:conf/icml/BruceDEPS0LMSAA24, DBLP:journals/corr/abs-2506-09985}.
Beyond static perception, several studies \citep{DBLP:conf/nips/DuY0DN0SA23, DBLP:conf/iclr/TianYZWL0P25, DBLP:conf/iclr/ZhangDLPW25} aim to model environment dynamics by predicting future observations, which can then be used for action selection via inverse dynamics prediction.
Recent work further adopts joint frameworks that generate future frames together with corresponding actions, either implicitly or explicitly, thereby improving temporal consistency and facilitating policy learning \citep{DBLP:journals/corr/abs-2508-10333, DBLP:conf/icml/abs-2412-14803, DBLP:journals/corr/abs-2507-04447, DBLP:journals/corr/abs-2509-06951}.
Building on these ideas, several methods propose unified models that are jointly trained on and leverage multiple modalities, enabling integrated understanding and generation across vision, language, and action spaces \citep{DBLP:journals/corr/abs-2506-21539, DBLP:journals/corr/abs-2512-00975, DBLP:journals/corr/abs-2512-09928, DBLP:journals/corr/abs-2506-19850, DBLP:conf/cvpr/ZhaoLKFZWLMHFHL25, DBLP:conf/icml/abs-2501-18867}.

\section{Methodology}
\label{sec:method}

\begin{figure*}[!t]
    \centering
    \includegraphics[width=\linewidth]{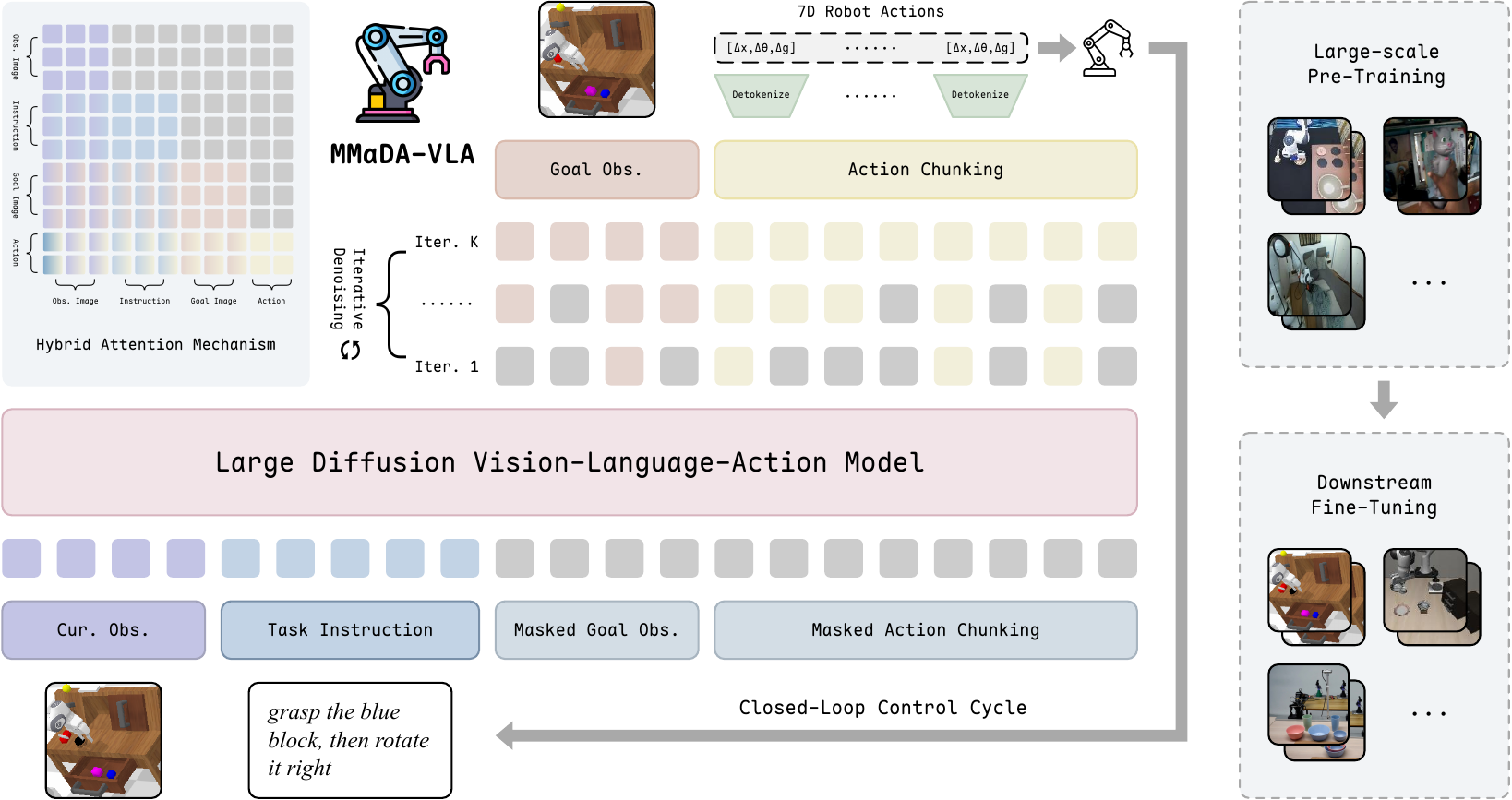}
    \caption{Schematic of the \model framework and training pipeline.}
    \Description{MMaDA-VLA combines current observations, task instructions, masked goal observations, and masked action chunks in a single large diffusion vision-language-action model. Iterative parallel denoising predicts a future goal observation and a chunk of seven-dimensional robot actions, followed by closed-loop execution. The model is first pre-trained on large-scale robot data and then fine-tuned on downstream tasks.}
    \label{fig:arch}
\end{figure*}

In this section, we present \model, a VLA model that unifies multi-modal instruction following and generation within a native discrete diffusion framework.
Figure~\ref{fig:arch} provides an overview of \model.

\subsection{Problem Formulation}

Conventional VLA approaches fine-tune a pretrained VLM on robotic manipulation datasets to learn a policy that predicts an action $a_t$ from the current visual observation $o_t$ and a language instruction $\ell$: $\hat{a}_t\sim\pi_{\theta}(a_t \mid o_t, \ell)$.
After the robot executes $a_t$, it receives an updated observation $o_{t+1}$, which is then used to infer the next action $a_{t+1}$.
This closed-loop process repeats until the task succeeds or a predefined maximum number of steps is reached, in which case the attempt is considered a failure.
We augment action generation with the parallel prediction of a goal visual observation:
\begin{equation}
    (\hat{o}_{t^{\prime}};\hat{a}_{t:t^{\prime}-1})\sim\pi_{\theta}(o_{t^{\prime}},a_{t:t^{\prime}-1}\mid o_{t},\ell),
\end{equation}
where $t^{\prime}=t+k$ and $k$ denotes the action chunk size. 
Specifically, we use the image-understanding and generation capabilities of a unified multi-modal model to introduce a world-model-like notion of dynamics into the VLA framework by training the model to predict the goal observation.
This additional objective encourages the policy to internalize task-relevant state evolution, thereby supporting more informed action generation.

\subsection{Architecture}

\paragraph{Data Tokenization.}
We adopt a unified discrete tokenization strategy across all modalities, enabling a single modeling framework that represents language, vision, and robot actions in a common token space.
With this representation, the model is trained with a single shared objective, masked-token prediction, regardless of modality.
Specifically, we use the LLaDA text tokenizer \citep{nie2025large}, the pretrained MAGVIT-v2 image quantizer \citep{DBLP:conf/iclr/YuLGVSMCGGHG0ER24} adopted from Show-o \citep{DBLP:conf/iclr/XieMBZWLGCYS25}, and an action tokenizer that discretizes each dimension of a continuous robot action into one of 256 bins \citep{DBLP:conf/corl/KimPKXB0RFSVKBT24}, with the bin width determined by the range of the training data.
Let $\tilde{o}_t$, $\tilde{\ell}$, and $\tilde{a}_t$ denote the token sequences obtained by applying the corresponding tokenizer $\tau$ to observation $o_t$, language $\ell$, and action $a_t$, respectively. For any modality input $x\in\{o_t,\ell,a_t\}$, we define its tokenized representation as $\tilde{x}=\tau(x)$.
All modalities are thus mapped to discrete tokens and represented in a shared vocabulary of size $|\mathcal{V}|$.
This unified representation supports a single learning objective across modalities, facilitates multi-modal fusion and cross-modal understanding, and improves scalability.

\paragraph{Multi-Modal Sequence Modeling.}
Based on this discrete-token formulation, we introduce a unified prompting strategy that standardizes the input--output format for VLA tasks and supports multi-modal instruction and generation:
\begin{equation}
    \resizebox{0.9\linewidth}{!}{$
        x=
        \underbrace{
            \texttt{[SOO]}\,\tilde{o}_{t}\,\texttt{[EOO]}\,\texttt{[SOL]}\,\tilde{\ell}\,\texttt{[EOL]}
        }_{\text{Instruction}}
        \underbrace{
            \texttt{[SOO]}\,\tilde{o}_{t^{\prime}}\,\texttt{[EOO]}\,\texttt{[SOA]}\,\tilde{a}_{t:t^{\prime}-1}\,\texttt{[EOA]}
        }_{\text{Generation}}
    $},
\end{equation}
where $\texttt{[SOX]}$ and $\texttt{[EOX]}$ are special tokens that mark the beginning and end of a modality sequence.
The masking patterns applied to $\tilde{o}_{t^{\prime}}$ and $\tilde{a}_{t:t^{\prime}-1}$ differ between training and inference.

\paragraph{Hybrid Attention Mechanism.}
To sufficiently exploit the capacity of large diffusion models, we propose a hybrid attention mechanism (top-left of Figure~\ref{fig:arch}).
Concretely, we apply bidirectional full attention for intra-modal interactions and causal attention for inter-modal interactions.
This arrangement enables global information exchange among tokens within the same modality while enforcing a directed flow of information across modalities.
Consequently, the two generation targets are effectively decoupled, yielding stronger and more stable representations for final action prediction.
Furthermore, autoregressive decoding is ill-suited to action tokens because the dimensions within each action chunk (e.g., a 7-DoF action vector) are inherently unordered and lack a meaningful sequential structure.
Therefore, imposing a generation order introduces an unnecessary inductive bias and exacerbates error accumulation as mistakes propagate to subsequent predictions.
In contrast, intra-modal full attention combined with parallel decoding better respects the unordered nature of action dimensions and mitigates compounding errors.
Moreover, iterative refinement allows action prediction to continually leverage intermediate features from goal-image generation, rather than depending only on the generated goal image \citep{DBLP:conf/icml/abs-2412-14803}.

\subsection{Learning Objective}
\label{sec:learn_obj}

During both pre-training and fine-tuning, we sample a diffusion step $s$ and randomly replace tokens in the generation part of the sequence $x$ with a special mask token \mask according to the corresponding cosine mask schedule \citep{DBLP:conf/cvpr/ChangZJLF22}.
Let $x^{(s)}\sim q_s(\cdot\mid x)$ denote the resulting masked sequence at diffusion step $s$, where $q_s$ is the corresponding forward corruption process.
The number of masked tokens is $N_s=\sum_{i=1}^{n}\mathbf{1}[x_{i}^{(s)}=\mask]$, where $n$ is the sequence length.

Under the unified framework, we adopt a single learning objective for the entire sequence across all modalities:
\begin{equation}
    \mathcal{L}(\theta)=-\mathbb{E}_{s,x,x^{(s)}\sim q_s(\cdot\mid x)}\left[\frac{1}{N_s}\sum_{i=1}^{n}\mathbf{1}[x_{i}^{(s)}=\mask]\log\pi_\theta(x_{i}\mid x^{(s)})\right].
\end{equation}


\subsection{Inference}
\label{sec:infer}

\paragraph{Iterative Denoising.}
At inference time, we replace all tokens in $\tilde{o}_{t'}$ and $\tilde{a}_{t:t'-1}$ with the mask token \mask and feed the resulting input $x^{(D)}$ into \model.
We then perform $D$ iterative denoising steps, analogous to diffusion, to recover these masked tokens and thereby predict the goal visual observation and the actions to be executed.
Specifically, \model performs $D$ discrete denoising steps, indexed by $d=D$ down to $1$, to produce a sequence of intermediate states.
We use $x^{(d)}$ to denote the intermediate state at step $d$, starting from $x^{(D)}$.
At each step $d$, \model $\pi_{\theta}$ estimates the distribution over the clean sequence, and the most likely sequence $\hat{x}^{(0)}$ is obtained via greedy decoding:
\begin{equation}
    \hat{x}^{(0)}=\arg\max_{\mathcal{V}}\!\;\pi_{\theta}(x^{(d)};\theta).
\end{equation}
We compute the number of tokens to mask according to the mask-scheduling function $\gamma$ as $\beta=\left\lceil \gamma\!\left(\frac{d}{D}\right)n^{\prime} \right\rceil$, where $n^{\prime}$ is the total number of tokens in $\tilde{o}_{t'}$ and $\tilde{a}_{t:t'-1}$.
A confidence-based remasking transition then yields $x^{(d-1)}$ by selectively updating the tokens in $x^{(d)}$ according to $\hat{x}^{(0)}$:
\begin{equation}
    x^{(d-1)}_{i}=
    \begin{cases}
    \!\mask, & \text{if}\ c_i < \mathrm{sort}([{c_{1},\dots,c_{n}}])[\beta] \\
    \hat{x}^{(0)}_{i}, & \text{otherwise},
    \end{cases}
\end{equation}
where $c_{i}$ is the confidence score for the $i$th token.
After the final denoising step, the final token sequence $\hat{x}^{(0)}$ is obtained.
We decode the different subsequences with their corresponding tokenizers to obtain $o_{t+1}$ and $a_t$.
The robotic arm executes $a_t$, transitioning to a new state, and the process repeats in a closed loop.

\newcommand{\liberonorline}[7]{#1 \citep{#2} & #3 & #4 & #5 & #6 & #7}
\newcommand{\liberotgxline}[7]{\tgx{#1 \citep{#2}} & \tgx{#3} & \tgx{#4} & \tgx{#5} & \tgx{#6} & \tgx{#7}}
\newcommand{\calvinnorline}[8]{#1 \citep{#2} & #3 & #4 & #5 & #6 & #7 & #8 \\}
\newcommand{\calvintgxline}[8]{\tgx{#1 \citep{#2}} & \tgx{#3} & \tgx{#4} & \tgx{#5} & \tgx{#6} & \tgx{#7} & \tgx{#8} \\}

\begin{table*}[!t]
\caption{Performance comparison on LIBERO \citep{DBLP:conf/nips/LiuZGFLZS23} and CALVIN \citep{DBLP:journals/ral/MeesHRB22}. For LIBERO, we report task success rates (\%); for CALVIN, we report subtask success rates (\%) and average sequence length. Continuous-action methods are shown in \grey{grey}.}
\label{tab:libero_calvin}
\centering
\small
\begin{tabular}{lccccclcccccc}
\toprule
\multicolumn{1}{c}{\multirow{2}{*}{\bf Method}} & \multicolumn{5}{c}{\bf LIBERO} & \multicolumn{1}{c}{\multirow{2}{*}{\bf Method}} & \multicolumn{6}{c}{\bf CALVIN} \\
\cmidrule(lr){2-6} \cmidrule(lr){8-13}
 & {\bf Spatial} & {\bf Object} & {\bf Goal} & {\bf Long} & {\bf Avg.} & & {\bf 1/5} & {\bf 2/5} & {\bf 3/5} & {\bf 4/5} & {\bf 5/5} & {\bf Avg. Len.} \\
\midrule
\liberotgxline{Diffusion Policy}{DBLP:conf/rss/ChiFDXCBS23}{78.3}{92.5}{68.3}{50.5}{72.4} & \calvintgxline{Diffusion Policy}{DBLP:conf/rss/ChiFDXCBS23}{40.2}{12.3}{2.6}{0.8}{0.0}{0.56}
\liberotgxline{Octo}{DBLP:conf/rss/GhoshWPBMDHK0LT24}{78.9}{85.7}{84.6}{51.1}{75.1} & \calvinnorline{RT-1}{DBLP:conf/rss/BrohanBCCDFGHHH23}{53.3}{22.2}{9.4}{3.8}{1.3}{0.90}
\liberonorline{OpenVLA}{DBLP:conf/corl/KimPKXB0RFSVKBT24}{84.9}{88.4}{79.2}{53.7}{76.5} & \calvinnorline{VLAS}{DBLP:conf/iclr/ZhaoD0GB0W25}{87.2}{64.2}{40.9}{28.1}{19.6}{2.40}
\liberonorline{SpatialVLA}{DBLP:journals/rss/abs-2501-15830}{88.2}{89.9}{78.6}{55.5}{78.1} & \calvintgxline{RoboFlamingo}{DBLP:conf/iclr/LiLZYXWCJ0LLK24}{82.4}{61.9}{46.6}{33.1}{23.5}{2.47}
\liberotgxline{DiT Policy}{DBLP:journals/corr/abs-2410-15959}{84.2}{96.3}{85.4}{63.8}{82.4} & \calvintgxline{Deer}{DBLP:conf/nips/YueWKHWSF024}{84.8}{72.3}{54.9}{44.6}{33.5}{2.90}
\liberonorline{$\pi_{0}$-FAST}{DBLP:journals/corr/abs-2501-09747}{96.4}{96.8}{88.6}{60.2}{85.5} & \calvinnorline{OpenVLA}{DBLP:conf/corl/KimPKXB0RFSVKBT24}{91.3}{77.8}{62.0}{52.1}{43.5}{3.27}
\liberotgxline{GR00T-N1}{DBLP:journals/corr/abs-2503-14734}{94.4}{97.6}{93.0}{90.6}{93.9} & \calvintgxline{$\pi_{0}$}{DBLP:journals/corr/abs-2410-24164}{93.8}{85.0}{76.7}{68.1}{59.9}{3.92}
\liberotgxline{$\pi_{0}$}{DBLP:journals/corr/abs-2410-24164}{96.8}{98.8}{95.8}{85.2}{94.2} & \calvinnorline{LLaDA-VLA}{DBLP:journals/corr/abs-2509-06932}{95.6}{87.7}{79.5}{73.9}{64.5}{4.01}
\liberonorline{Discrete Diffusion VLA}{DBLP:journals/corr/abs-2508-20072}{97.2}{98.6}{97.4}{92.0}{96.3} & \calvintgxline{OpenHelix}{DBLP:journals/corr/abs-2505-03912}{97.1}{91.4}{82.8}{72.6}{64.1}{4.08}
\liberotgxline{$\pi_{0.5}$}{DBLP:journals/corr/abs-2504-16054}{98.8}{98.2}{98.0}{92.4}{96.8} & \calvintgxline{OpenVLA-OFT}{DBLP:journals/corr/abs-2502-19645}{96.3}{89.1}{82.4}{75.8}{66.5}{4.10}
\liberotgxline{OpenVLA-OFT}{DBLP:journals/corr/abs-2502-19645}{97.6}{98.4}{97.9}{94.5}{97.1} & \calvintgxline{RoboVLMs}{DBLP:journals/corr/abs-2412-14058}{98.0}{93.6}{85.4}{77.8}{70.4}{4.25}
\liberotgxline{VLA-Adapter}{DBLP:journals/corr/abs-2509-09372}{97.8}{99.2}{97.2}{95.0}{97.3} & \calvintgxline{VLA-Adapter}{DBLP:journals/corr/abs-2509-09372}{99.1}{94.6}{88.8}{82.8}{76.5}{4.42}
\midrule
\rowcolor[rgb]{.949,.949,.949}
\multicolumn{1}{l}{\quad\textit{World Model}} & \multicolumn{5}{c}{} & \multicolumn{1}{l}{\quad\textit{World Model}} & \multicolumn{6}{c}{} \\
\liberonorline{WorldVLA}{DBLP:journals/corr/abs-2506-21539}{85.6}{89.0}{82.6}{59.0}{79.1} & \calvintgxline{SuSIE}{DBLP:conf/iclr/BlackNAWFKL24}{87.0}{69.0}{49.0}{38.0}{26.0}{2.69}
\liberonorline{CoT-VLA}{DBLP:conf/cvpr/ZhaoLKFZWLMHFHL25}{87.5}{91.6}{87.6}{69.0}{81.1} & \calvintgxline{GEVRM}{DBLP:conf/iclr/ZhangDLPW25}{92.0}{70.0}{54.0}{41.0}{26.0}{2.83}
\liberonorline{FlowVLA}{DBLP:journals/corr/abs-2508.18269}{93.2}{95.0}{91.6}{72.6}{88.1} & \calvintgxline{GR-1}{DBLP:conf/iclr/WuJCCXLLLK24}{85.4}{71.2}{59.6}{49.7}{40.1}{3.06}
\liberotgxline{Seer}{DBLP:conf/iclr/TianYZWL0P25}{\na}{\na}{\na}{87.7}{87.7} & \calvinnorline{ReconVLA}{DBLP:journals/corr/abs-2508-10333}{95.6}{87.6}{76.9}{69.3}{64.1}{3.95}
\liberotgxline{DreamVLA}{DBLP:journals/corr/abs-2507-04447}{97.5}{94.0}{89.5}{89.5}{92.6} & \calvinnorline{UP-VLA}{DBLP:conf/icml/abs-2501-18867}{92.8}{86.5}{81.5}{76.9}{69.9}{4.08}
\liberonorline{UD-VLA}{DBLP:journals/corr/abs-2511-01718}{94.1}{95.7}{91.2}{89.6}{92.7} & \calvintgxline{Seer}{DBLP:conf/iclr/TianYZWL0P25}{96.3}{91.6}{86.1}{80.3}{74.0}{4.28}
\liberonorline{RynnVLA-002-Discrete}{cen2025rynnvla002unifiedvisionlanguageactionworld}{94.2}{96.8}{94.6}{87.6}{93.3} & \calvintgxline{VPP}{DBLP:conf/icml/abs-2412-14803}{95.7}{91.2}{86.3}{81.0}{75.0}{4.33}
\liberonorline{UniVLA}{DBLP:journals/corr/abs-2506-19850}{95.4}{98.8}{93.5}{94.0}{95.5} & \calvinnorline{UniVLA}{DBLP:journals/corr/abs-2506-19850}{98.9}{94.8}{89.0}{82.8}{75.1}{4.41}
\liberotgxline{$\mathcal{F}_1$}{DBLP:journals/corr/abs-2509-06951}{98.2}{97.8}{95.4}{91.3}{95.7} & \calvintgxline{HiF-VLA}{DBLP:journals/corr/abs-2512-09928}{98.5}{94.1}{88.1}{81.4}{73.1}{4.35}
\liberonorline{MM-ACT}{DBLP:journals/corr/abs-2512-00975}{97.8}{99.4}{94.8}{93.0}{96.3} & \calvintgxline{DreamVLA}{DBLP:journals/corr/abs-2507-04447}{98.2}{94.6}{89.5}{83.4}{78.1}{4.44}

\rowcolor{lavendermist} {\bf \model (Ours)} & {\bf 98.8} & {\bf 99.8} & {\bf 98.0} & {\bf 95.2} & {\bf 98.0} & {\bf \model (Ours)} & {\bf 99.8} & {\bf 98.6} & {\bf 96.3} & {\bf 93.5} & {\bf 89.7} & {\bf 4.78} \\
\bottomrule
\end{tabular}
\end{table*}

\paragraph{Key-Value Cache.}
While iterative denoising offers flexible generation, it suffers from high inference latency because similar computations are repeatedly performed across denoising steps.
To meet the stringent real-time requirements of robot manipulation, we adopt a training-free caching framework to improve inference efficiency.
Notably, the instruction part remains fixed throughout denoising, so its intermediate representations are stable and can be cached over long horizons.
In contrast, the generation part evolves across steps; however, this evolution is typically sparse, with only a small subset of generated tokens changing substantially at each step \citep{DBLP:journals/corr/abs-2506-06295}.
Accordingly, for each Transformer layer $l$, we cache the key $\mathrm{K}_l$, value $\mathrm{V}_l$, attention output $\mathrm{AttnOut}_l$, and feed-forward network output $\mathrm{FFNOut}_l$ in a cache $\mathcal{C}$.
The cache is refreshed every $\lambda$ denoising steps.
For the generation part, we perform selective refresh by updating only the $\lfloor\rho n^{\prime}\rfloor$ tokens whose current and cached value vectors have the lowest cosine similarity, where $\rho$ denotes the adaptive update ratio.

\section{Experiments}
\label{sec:exp}

\subsection{Evaluation}
\label{sec:eval}

We evaluate \model on the LIBERO \citep{DBLP:conf/nips/LiuZGFLZS23} and CALVIN \citep{DBLP:journals/ral/MeesHRB22} simulation benchmarks, as well as on real-world tasks, to assess multi-task generalization and long-horizon performance.

\paragraph{LIBERO Simulation Benchmark.}
The LIBERO benchmark \citep{DBLP:conf/nips/LiuZGFLZS23} is a comprehensive suite for evaluating multi-task and lifelong robotic manipulation.
It comprises four task suites designed to probe distinct forms of generalization.
Spatial evaluates spatial reasoning by varying scene layouts while keeping the set of objects fixed.
Object tests object-level generalization by varying the objects within an otherwise fixed scene.
Goal assesses goal-conditioned behavior by changing task goals under a consistent environment configuration.
Finally, Long contains compositional, long-horizon tasks with diverse objects, layouts, and goals, thereby stressing temporal consistency and compositional reasoning.

\paragraph{CALVIN Simulation Benchmark.}
The CALVIN benchmark \citep{DBLP:journals/ral/MeesHRB22} evaluates long-horizon, language-conditioned robotic manipulation.
It comprises four simulated environments (A, B, C, and D), each providing demonstration trajectories collected via human teleoperation.
The CALVIN long-horizon challenge consists of sequences of five subtasks to be executed consecutively.
Performance is quantified by per-subtask success rates and the average number of completed subtasks, computed over 1,000 rollout episodes to ensure a fair comparison.
In the ABC$\rightarrow$D setting, models are trained on environments A, B, and C and evaluated on environment D to assess environmental generalization in addition to long-horizon execution. The test environment differs in table textures, furniture positions, and color schemes.

\begin{figure*}[!t]
    \centering
    \includegraphics[width=\linewidth]{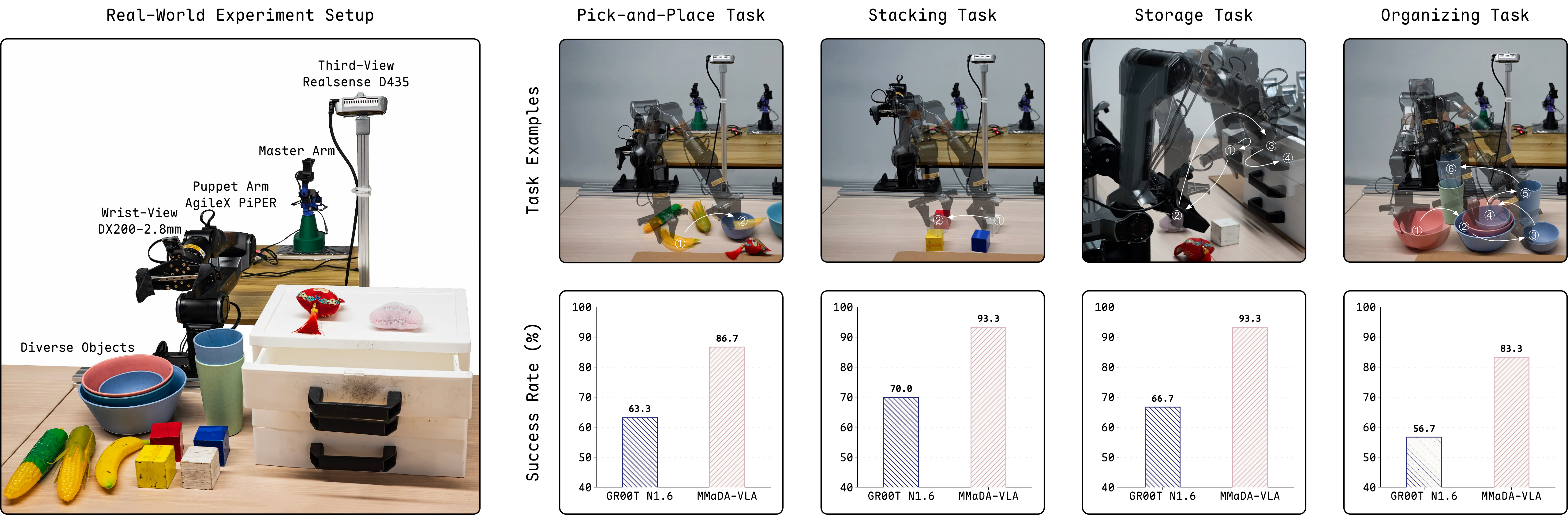}
    \caption{Real-world experiment setup, task examples, and performance comparison.}
    \Description{The experiment uses an AgileX PiPER arm with wrist-view and third-person cameras and diverse tabletop objects. Four task columns show pick-and-place, block stacking, drawer storage, and bowl-and-cup organization. Bar charts compare MMaDA-VLA with GR00T N1.6; MMaDA-VLA reaches 86.7, 93.3, 93.3, and 83.3 percent success, respectively, versus 63.3, 70.0, 66.7, and 56.7 percent.}
    \label{fig:real}
\end{figure*}

\paragraph{Real-World Evaluation.}

\newcommand{\dataline}[4]{#1 \citep{#2} & #3 & #4}

\begin{table}[!t]
\caption{Dataset mixture details for \model.}
\label{tab:data_mix}
\centering
\small
\resizebox{\linewidth}{!}{
\begin{tabular}{lcclcc}
\toprule
\multicolumn{1}{c}{\bf Dataset} & {\bf Ratio (\%)} & {\bf Samples} & \multicolumn{1}{c}{\bf Dataset} & {\bf Ratio (\%)} & {\bf Samples} \\
\midrule
\dataline{DROID}{DBLP:conf/rss/KhazatskyP0BDKN24}{49.94}{26675496} &
\dataline{Austin Sirius Dataset}{DBLP:conf/rss/LiuNZBZ23}{0.52}{277701} \\
\dataline{BC-Z}{DBLP:conf/corl/JangIKKELLF21}{9.95}{5314594} &
\dataline{Taco Play}{DBLP:conf/corl/Rosete-BeasMKBB22,DBLP:conf/icra/MeesBB23}{0.38}{201002} \\
\dataline{Language Table}{DBLP:journals/ral/Corey23}{9.88}{5276570} &
\dataline{Roboturk}{DBLP:conf/corl/MandlekarZGBSTG18}{0.30}{161268} \\
\dataline{Furniture Bench Dataset}{DBLP:conf/rss/HeoLLL23}{7.35}{3927655} &
\dataline{IAMLab CMU Pickup Insert}{DBLP:conf/corl/SaxenaSK23}{0.27}{143715} \\
\dataline{Fractal}{DBLP:conf/rss/BrohanBCCDFGHHH23}{6.44}{3438450} &
\dataline{Berkeley Autolab UR5}{BerkeleyUR5Website}{0.16}{84197} \\
\dataline{Bridge V2}{DBLP:conf/corl/WalkeBZVZHHMKDL23}{3.15}{1680298} &
\dataline{Viola}{DBLP:journals/corr/abs-2210-11339}{0.13}{68371} \\
\dataline{Kuka}{DBLP:journals/corr/abs-1806-10293}{3.03}{1616454} &
\dataline{Jaco Play}{dass2023jacoplay}{0.12}{66221} \\
\dataline{FMB Dataset}{DBLP:journals/ijrr/LuoXLTLWAL25}{2.06}{1103009} &
\dataline{Berkeley Fanuc Manipulation}{fanuc_manipulation2023}{0.11}{60951} \\
\dataline{CALVIN}{DBLP:journals/ral/MeesHRB22}{1.87}{1000263} &
\dataline{Austin Buds Dataset}{DBLP:journals/ral/ZhuSZ22}{0.06}{33910} \\
\dataline{LIBERO}{DBLP:conf/nips/LiuZGFLZS23}{1.55}{829416} &
\dataline{NYU Franka Play Dataset}{DBLP:conf/iclr/CuiWSP23}{0.06}{32986} \\
\dataline{Stanford Hydra Dataset}{DBLP:conf/corl/BelkhaleCS23}{0.67}{355952} &
\dataline{Berkeley Cable Routing}{DBLP:journals/corr/abs-2307-08927}{0.06}{32385} \\
\dataline{UTAustin Mutex}{DBLP:conf/corl/ShahMZ23}{0.67}{355881} &
\dataline{CMU Stretch}{DBLP:conf/rss/MendoncaBP23}{0.05}{24474} \\
\dataline{Austin Sailor Dataset}{DBLP:conf/corl/NasirianyGMZ22}{0.66}{352132} &
\dataline{DLR EDAN Shared Control}{DBLP:conf/icra/QuereHIBLSV20}{0.02}{8510} \\
\dataline{Toto}{DBLP:conf/icra/ZhouDSRPHJYAPFG23}{0.54}{290529} &
\dataline{UCSD Kitchen Dataset}{ucsd_kitchens}{0.01}{3368} \\
\bottomrule
\end{tabular}
}
\end{table}

As shown in Figure~\ref{fig:real}, we conduct real-world evaluations on an AgileX PiPER 6-DoF robotic arm equipped with a 1-DoF gripper.
A RealSense D435 provides third-person observations, while a DX200-2.8mm camera mounted on the gripper provides wrist-view observations.
We evaluate \model across four experimental categories:
(1) Simple pick-and-place task: the robot must pick up a specified object and place it into a designated container.
To assess robustness to environmental changes, we introduce disturbances during execution, including visually similar distractor objects (e.g., a banana and an ear of corn) and manual displacement of the target container.
This setting evaluates the model's ability to ground natural-language semantics in visual observations and react to dynamic scene variations.
(2) Precision stacking task: the robot is instructed to stack a block of a specified color onto another block of a specified color.
This task requires accurate semantic understanding, precise grasping, and fine alignment during placement.
(3) Complex storage task: the robot must open a drawer, pick up a specified object, place it inside the drawer, and then close the drawer.
Beyond standard pick-and-place operations, this task involves diverse drawer interactions (e.g., pulling, pushing, and rotational motions).
The object set includes both rigid blocks and soft plush toys, assessing robustness to diverse object properties and grasping behaviors.
(4) Long-horizon organizing task: the robot must organize tableware on the tabletop, including two cups and three bowls.
This long-horizon task requires coordinating multiple sequential manipulations, including repeated stacking and placement operations.
The irregular geometries of the cups and bowls further challenge the model's ability to handle varied grasping and manipulation strategies.
Overall, we collect 300 demonstrations for each task to fine-tune the models.
For evaluation, we run 30 trials per task and report the task success rate (\%).

\subsection{Implementation Details}

\begin{table}[!t]
\caption{Hyperparameter settings. $^{\sharp}$ and $^{\dagger}$ denote the settings used for LIBERO and CALVIN, respectively.}
\label{tab:params}
\centering
\small
\resizebox{\linewidth}{!}{
\begin{tabular}{lcclcc}
\toprule
\multicolumn{1}{c}{\bf Configuration} & \multicolumn{1}{c}{\bf Pre-Training} & \multicolumn{1}{c}{\bf Fine-Tuning} & \multicolumn{1}{c}{\bf Configuration} & \multicolumn{1}{c}{\bf Pre-Training} & \multicolumn{1}{c}{\bf Fine-Tuning} \\
\midrule
Backbone & \multicolumn{2}{c}{MMaDA-8B-Base \cite{yang2025mmada}} & Image Tokenizer & \multicolumn{2}{c}{MAGVIT-v2 \cite{DBLP:conf/iclr/YuLGVSMCGGHG0ER24}} \\
Global Batch Size & 640 & 80$^{\sharp}$ \& 320$^{\dagger}$ & Numerical Precision & \multicolumn{2}{c}{BFloat16} \\
Optimizer & \multicolumn{2}{c}{AdamW \cite{DBLP:conf/iclr/LoshchilovH19}} & Learning Rate & \multicolumn{2}{c}{$1\times10^{-4}$} \\
Epochs & 1 & 20$\sim$40$^{\sharp}$ \& 2$^{\dagger}$ & Weight Decay & \multicolumn{2}{c}{0.01} \\
Learning-Rate Schedule & \multicolumn{2}{c}{Cosine Decay} & Warm-up Ratio & \multicolumn{2}{c}{0.01} \\
Textual Instruction Length & \multicolumn{2}{c}{128} & Action Chunk Size & 5 & 5$^{\sharp}$ \& 10$^{\dagger}$ \\
\bottomrule
\end{tabular}
}
\end{table}

\model uses MMaDA-8B-Base \citep{yang2025mmada}, a unified multi-modal diffusion language model, as its 8-billion-parameter backbone.
RGB observation images from both the third-person and wrist-view cameras are vertically concatenated and resized to a resolution of $256\times256$.
The proprioceptive state of the robot is appended, in textual form, after the task instruction.
In the pre-training stage, we leverage existing large-scale, cross-embodiment robotic manipulation datasets to learn general relationships between visual observations and manipulation actions.
We collect approximately $53$ million training samples for pre-training; dataset mixture details are provided in Table~\ref{tab:data_mix}.
To ensure a fair and consistent evaluation under the ABC$\rightarrow$D setting, we exclude the Part D data from CALVIN.
In the fine-tuning stage, we set the action chunk size to $5$ for LIBERO \citep{DBLP:conf/nips/LiuZGFLZS23} and $10$ for long-horizon CALVIN \citep{DBLP:journals/ral/MeesHRB22}.
Leveraging DeepSpeed \citep{DBLP:conf/kdd/RasleyRRH20, DBLP:conf/sc/RajbhandariRRH20}, pre-training \model requires approximately $30$ hours on eight nodes, each equipped with eight NVIDIA H800 GPUs with 80~GB of memory per GPU.
Detailed configurations are provided in Table~\ref{tab:params}.

\subsection{Main Results}

\begin{figure*}[!t]
    \centering
    \includegraphics[width=\linewidth]{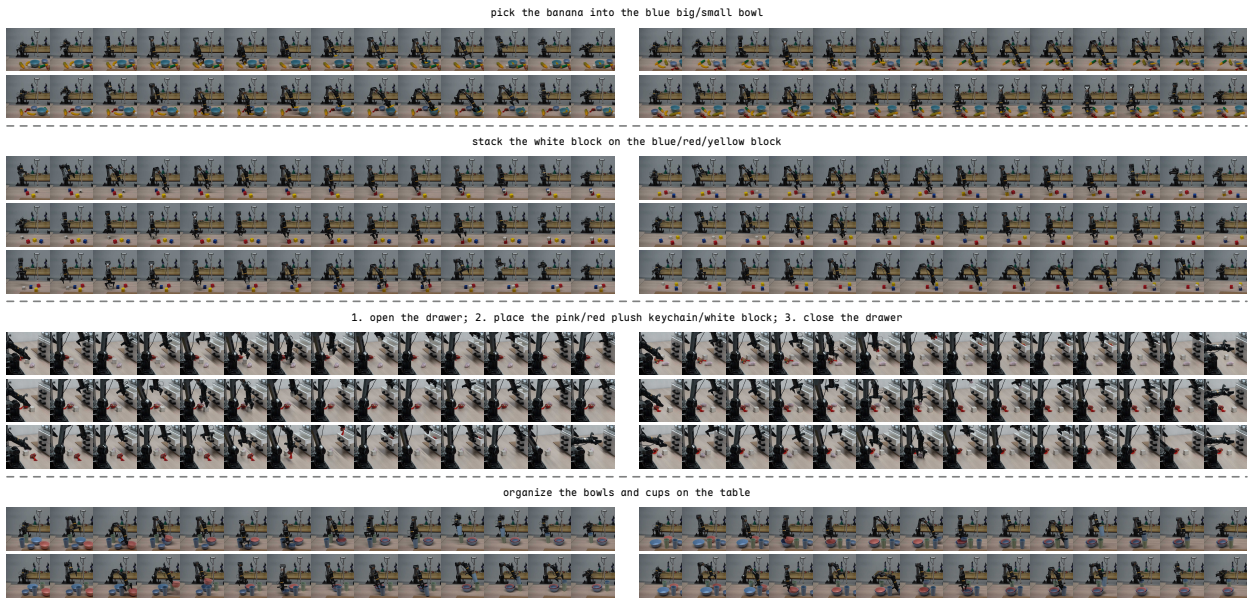}
    \caption{\model execution examples on real-world tasks.}
    \Description{Time-ordered image sequences show MMaDA-VLA executing four real-world tasks: placing a banana into a bowl, stacking colored blocks, opening a drawer to store an object and then closing it, and organizing bowls and cups on a table.}
    \label{fig:vis_real}
\end{figure*}

Table~\ref{tab:libero_calvin} and Figure~\ref{fig:real} show that \model achieves the highest reported performance among the compared methods on both simulation benchmarks and all four real-world tasks.

\paragraph{LIBERO}

On LIBERO, \model achieves a 98.0\% average success rate, surpassing the strongest reported baseline, VLA-Adapter \citep{DBLP:journals/corr/abs-2509-09372} (97.3\%), by 0.7 percentage points.
This gain is particularly notable because visual-prediction-based VLAs generally trail vanilla VLAs on this benchmark, indicating that our unified discrete modeling improves the generalization of goal prediction and, consequently, action generation.

\paragraph{CALVIN}

On CALVIN, \model reaches an average successful execution length of 4.78, improving over the strongest reported baseline, DreamVLA \citep{DBLP:journals/corr/abs-2507-04447} (4.44), by 0.34.
It further achieves an 89.7\% success rate on the fifth task, 11.6 percentage points above DreamVLA's 78.1\%.
These gains demonstrate improved long-horizon consistency and robustness across extended action sequences.

\paragraph{Real-World.}

In Figure~\ref{fig:real}, \model achieves success rates of 83.3\%--93.3\% across four real-world tasks, outperforming GR00T N1.6 \citep{DBLP:journals/corr/abs-2503-14734} by 23.3--26.6 percentage points.
Failures mainly involve imprecise grasps at narrow openings, insufficient drawer displacement, and unstable stacking of tall cups; nevertheless, the model can often recover through corrective actions.
Additional examples are provided in Figure~\ref{fig:vis_real}.

\begin{table}[!t]
\caption{Ablation studies on CALVIN.}
\label{tab:abla}
\centering
\small
\begin{tabular}{lc}
\toprule
\multicolumn{1}{c}{\bf Method} & {\bf Average Length} \\
\midrule
\rowcolor[rgb]{ .949,  .949,  .949} {\bf \model} (w/o Pre-Training) & {\bf 4.56} \\
\quad w/o World-Model & 4.08 \\
\quad w/o Parallel Denoising & 4.38 \\
\quad w/ Causal Attention & 4.49 \\
\quad w/ Bidirectional Attention & 4.52 \\
\bottomrule
\end{tabular}
\end{table}

\section{Analysis}
\label{sec:ana}

\subsection{Ablation Studies}

As summarized in Table~\ref{tab:abla}, we perform a series of ablation studies on the CALVIN benchmark \citep{DBLP:journals/ral/MeesHRB22}.
To avoid confounding effects from pre-training and to reduce computational cost, all ablations are evaluated without pre-training.
We compare \model fine-tuned from scratch against the following variants:
(1) \model w/o World-Model, which predicts robot actions without goal image prediction;
(2) \model w/o Parallel Denoising, which first generates the complete goal image and then predicts actions;
(3) \model w/ Causal Attention, which replaces hybrid attention with standard causal attention commonly used in autoregressive models;
and (4) \model w/ Bidirectional Attention, which instead uses full bidirectional attention as in BERT \citep{DBLP:conf/naacl/DevlinCLT19}.

Replacing the attention mechanism has a smaller effect on performance than modifying the overall generative paradigm.
\model w/o World-Model can be viewed as a vanilla VLA model built on a discrete diffusion framework.
Removing environment-dynamics modeling reduces the average execution length by 0.48.
Without Parallel Denoising, generating future observations before actions prevents action prediction from exploiting intermediate hidden states produced during image generation.
In addition, deterministically predicted goals can introduce cumulative errors that propagate to action generation, resulting in a decrease of 0.18.
By contrast, parallel denoising allows the model to leverage all tokens determined thus far at each iteration to better infer the remaining tokens.
Regarding attention mechanisms, both causal attention and fully bidirectional attention cause only minor performance decreases of less than 0.1.
Causal attention limits information exchange within each modality, whereas fully bidirectional attention can introduce noise through cross-modal information leakage.
Overall, \model achieves the highest average execution length of 4.56 by combining parallel denoising with hybrid attention.

\begin{table}[!t]
\caption{Effectiveness of pre-training.}
\label{tab:pretrain}
\centering
\small
\resizebox{0.85\linewidth}{!}{
\begin{tabular}{lccc}
\toprule
\multicolumn{1}{c}{\multirow{2}{*}{\bf Method}} & {\bf LIBERO} & {\bf CALVIN} \\
\cmidrule(lr){2-2} \cmidrule(lr){3-3}
 & {\bf Average Success Rate (\%)} & {\bf Average Sequence Length} \\
\midrule
\rowcolor[rgb]{ .949,  .949,  .949} {\bf \model} & {\bf 98.0} & {\bf 4.78} \\
\quad w/o Pre-Training & 94.5 & 4.56 \\
\bottomrule
\end{tabular}
}
\end{table}

\begin{figure*}[!t]
    \centering
    \includegraphics[width=\linewidth]{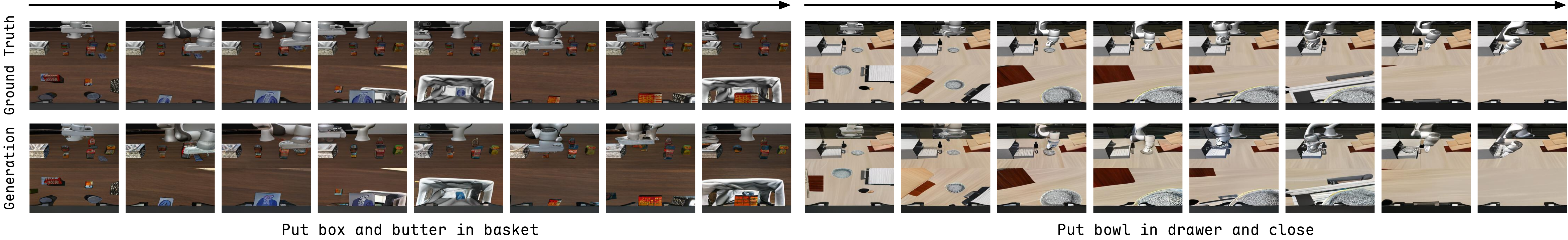}
    \caption{Visualization of generated goal observations versus ground-truth observations. For each timestep, we form a single composite frame by vertically concatenating the third-person view (top) and the wrist-mounted view (bottom).}
    \Description{Two task sequences compare ground-truth observations in the top row with generated observations in the bottom row. Across the tasks of putting a box and butter in a basket and putting a bowl in a drawer before closing it, the generated sequences preserve the main object motion and task progression while losing some fine visual detail.}
    \label{fig:vis_gen}
\end{figure*}

\begin{figure}[!t]
  \centering
  \begin{subfigure}[t]{0.43\linewidth}
    \includegraphics[width=\linewidth]{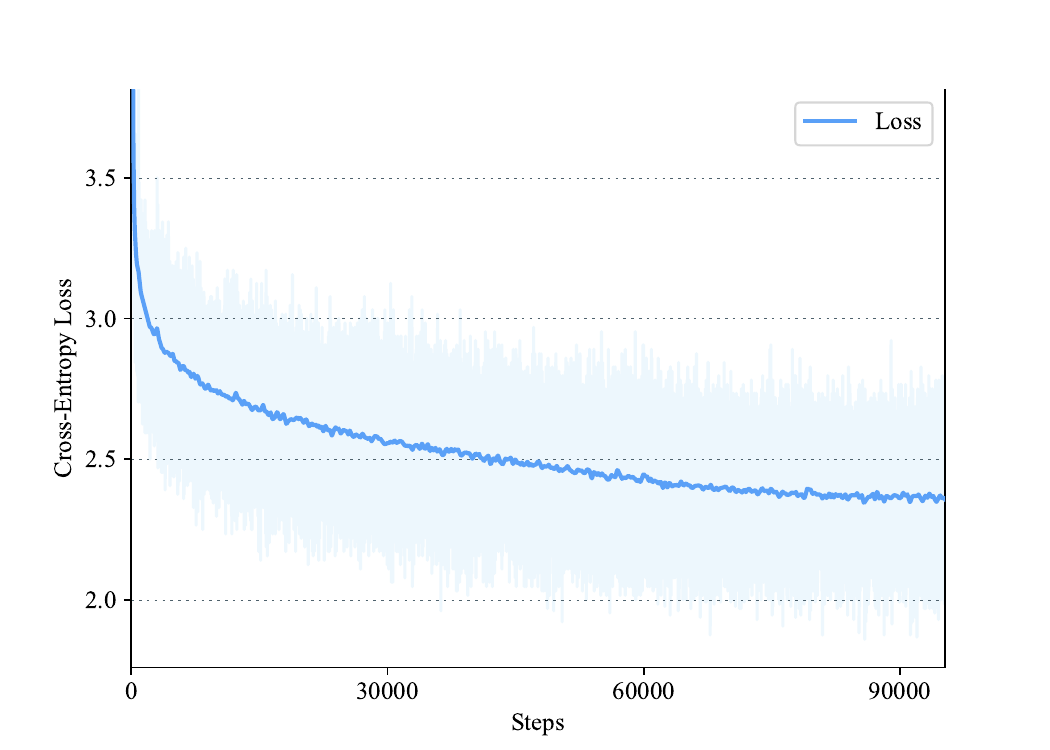}
    \caption{Loss curve.}
    \label{fig:loss}
  \end{subfigure}
  \hfill
  \begin{subfigure}[t]{0.43\linewidth}
    \includegraphics[width=\linewidth]{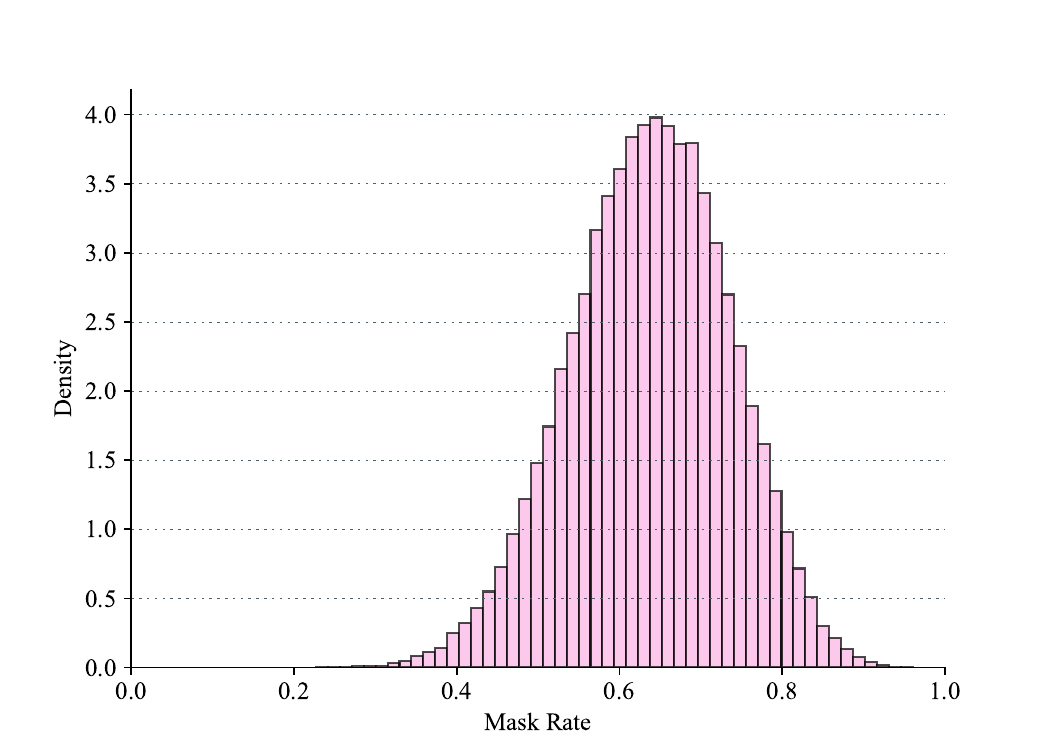}
    \caption{Mask rate density.}
    \label{fig:maskrate}
  \end{subfigure}
  \caption{Pre-training visualization.}
  \Description{The left plot shows cross-entropy loss varying within a bounded range over roughly 110,000 pre-training steps. The right plot shows the sampled mask-rate density over values from zero to one.}
  \label{fig:vis_pre}
\end{figure}

\subsection{Pre-Training}

Table~\ref{tab:pretrain} compares the performance of \model with and without large-scale pre-training.
Pre-training yields consistent improvements on both benchmarks.
On LIBERO, it increases the multi-task generalization score from 94.5\% to 98.0\%, a gain of 3.5 percentage points.
On CALVIN, it improves the average execution length on long-horizon tasks from 4.56 to 4.78, corresponding to an absolute increase of 0.22.
Thus, pre-training improves model generalization and execution accuracy by leveraging large-scale, diverse data.
Furthermore, we visualize the pre-training process using the loss curve and mask-rate statistics in Figure~\ref{fig:vis_pre}.
Because the mask rate is sampled according to the cosine schedule described in Section~\ref{sec:learn_obj}, the training loss varies within a bounded range that correlates with the sampled mask ratio.
The smoothed loss curve indicates that the model initially adapts rapidly to the newly introduced action tokens and subsequently learns cross-embodiment robot manipulation skills from large-scale data and egocentric human demonstrations. This progression improves generalization.
Moreover, the broad coverage of sampled mask rates supports stable learning across different levels of masking, which is important for each step of the iterative multi-step inference procedure.

\subsection{Visual Generation}

To assess both the strengths and limitations of the world-model behavior induced by visual generation, we visualize representative predictions of future observations in Figure~\ref{fig:vis_gen}.
Across a variety of environments and object appearances, the generated goal images are generally consistent with the instructions and preserve the high-level task dynamics, remaining broadly aligned with the corresponding ground-truth trajectories.
This high level of consistency suggests that the model's visual generation generalizes across diverse environments and can support the modeling of environment dynamics.
At the same time, the predicted images exhibit limited visual fidelity.
Fine-grained elements, such as the gripper geometry and small or visually complex objects, are often blurred or rendered inaccurately, as illustrated by the end effector in the left example.
This degradation is expected because we use a compact image representation with a small number of tokens for computational efficiency.
Despite the reduced pixel-level accuracy, the generated frames still convey task progression reliably and provide useful anticipatory cues for downstream control and action planning.

\begin{figure}[!t]
    \centering
    \includegraphics[width=0.8\linewidth]{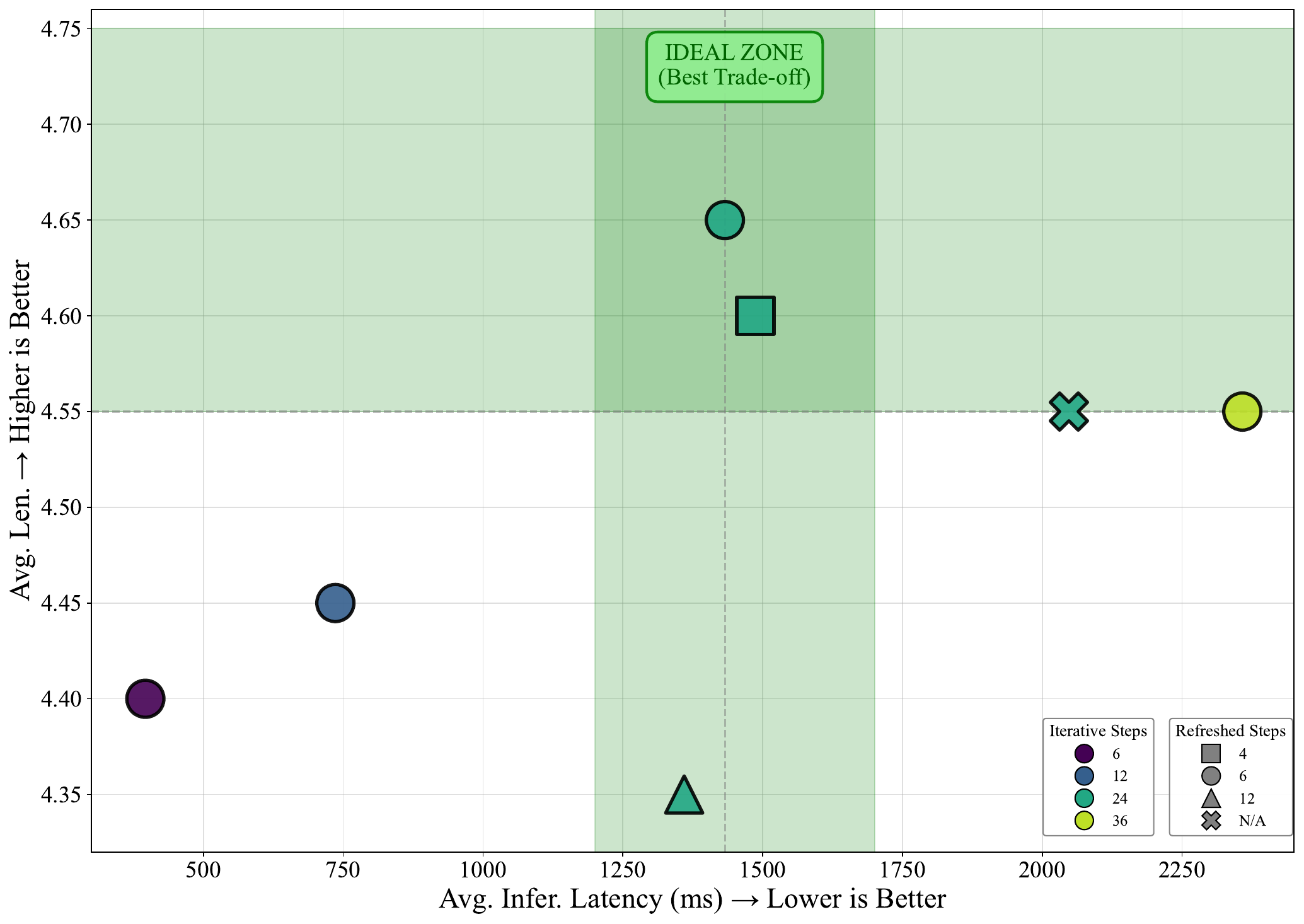}
    \caption{Inference latency and performance under different iterative denoising and cache-refresh settings.}
    \Description{A scatter plot compares average CALVIN task length against inference latency under different iterative-denoising and cache-refresh settings. The best trade-off region is around 1.4 to 1.5 seconds of latency with an average length above 4.55; 24 iterative steps achieve the highest average length of about 4.65.}
    \label{fig:infer_latency}
\end{figure}

\subsection{Efficiency}

We investigate the model's average inference latency and performance on a single NVIDIA A100 GPU under varying numbers of iterative denoising and cache-refresh steps, as illustrated in Figure~\ref{fig:infer_latency}.
We evaluate all configurations on the same set of 20 tasks randomly sampled from the CALVIN benchmark.
Under identical settings, the key-value cache improves inference speed, reducing average latency from 2.0~s to 1.4~s.
As the number of iterative denoising steps increases, performance initially improves but eventually declines.
This trend indicates that too few steps may cause premature commitment to erroneous tokens, whereas too many steps may result in repeated revisions to the token sequence.
The model achieves its best performance with 24 denoising steps, yielding an average length of 4.65.
Increasing the number of cache-refresh steps slightly improves inference speed but also introduces considerable performance fluctuations.
Based on these findings, we select 24 denoising steps and 6 cache-refresh steps to balance inference speed and model performance.

\section{Conclusion}
\label{sec:con}

We present \model, a fully native discrete diffusion VLA that unifies multimodal understanding and generation.
Through iterative denoising, \model jointly refines goal observations and action tokens within a unified training and inference framework.
Pre-trained on large-scale cross-embodiment data and fine-tuned on downstream tasks, \model achieves state-of-the-art performance on simulation benchmarks and real-world tasks, demonstrating accurate execution and strong generalization.

\begin{acks}
This work was supported by the Brain Science and Brain-like Intelligence Technology---National Science and Technology Major Project (Grant No. 2022ZD0208800).
\end{acks}

\bibliographystyle{ACM-Reference-Format}
\bibliography{main}

@inproceedings{DBLP:conf/rss/BrohanBCCDFGHHH23,
 author = {Anthony Brohan and others},
 booktitle = {Proc. of RSS},
 title = {{R}{T}-1: {R}obotics {T}ransformer for {R}eal-{W}orld {C}ontrol at {S}cale},
 year = {2023}
}

@inproceedings{DBLP:conf/corl/KimPKXB0RFSVKBT24,
 author = {Moo Jin Kim and others},
 booktitle = {Conference on Robot Learning, 6-9 November 2024, Munich, Germany},
 pages = {2679--2713},
 title = {{O}pen{V}{L}{A}: {A}n {O}pen-{S}ource {V}ision-{L}anguage-{A}ction {M}odel},
 year = {2024}
}

@inproceedings{DBLP:conf/iclr/BlackNAWFKL24,
 author = {Kevin Black and others},
 booktitle = {Proc. of ICLR},
 title = {{Z}ero-{S}hot {R}obotic {M}anipulation with {P}re-{T}rained {I}mage-{E}diting {D}iffusion
{M}odels},
 year = {2024}
}

@inproceedings{DBLP:conf/iclr/WuJCCXLLLK24,
 author = {Hongtao Wu and others},
 booktitle = {Proc. of ICLR},
 title = {{U}nleashing {L}arge-{S}cale {V}ideo {G}enerative {P}re-training for {V}isual {R}obot
{M}anipulation},
 year = {2024}
}

@inproceedings{DBLP:conf/iclr/ZhangDLPW25,
 author = {Hongyin Zhang and others},
 booktitle = {Proc. of ICLR},
 title = {{G}{E}{V}{R}{M}: {G}oal-{E}xpressive {V}ideo {G}eneration {M}odel {F}or {R}obust {V}isual
{M}anipulation},
 year = {2025}
}

@inproceedings{DBLP:conf/nips/DuY0DN0SA23,
 author = {Yilun Du and others},
 booktitle = {Proc. of NeurIPS},
 title = {{L}earning {U}niversal {P}olicies via {T}ext-{G}uided {V}ideo {G}eneration},
 year = {2023}
}

@inproceedings{DBLP:conf/icml/abs-2412-14803,
 author = {Yucheng Hu and others},
 booktitle = {Proc. of ICML},
 title = {{V}ideo {P}rediction {P}olicy: {A} {G}eneralist {R}obot {P}olicy with {P}redictive
{V}isual {R}epresentations},
 year = {2025}
}

@inproceedings{DBLP:conf/iclr/LiLZYXWCJ0LLK24,
 author = {Xinghang Li and others},
 booktitle = {Proc. of ICLR},
 title = {{V}ision-{L}anguage {F}oundation {M}odels as {E}ffective {R}obot {I}mitators},
 year = {2024}
}

@inproceedings{DBLP:conf/rss/ChiFDXCBS23,
 author = {Cheng Chi and others},
 booktitle = {Proc. of RSS},
 title = {{D}iffusion {P}olicy: {V}isuomotor {P}olicy {L}earning via {A}ction {D}iffusion},
 year = {2023}
}

@inproceedings{DBLP:conf/icml/abs-2501-18867,
 author = {Jianke Zhang and others},
 booktitle = {Proc. of ICML},
 title = {{U}{P}-{V}{L}{A}: {A} {U}nified {U}nderstanding and {P}rediction {M}odel for {E}mbodied
{A}gent},
 year = {2025}
}

@inproceedings{DBLP:conf/nips/LiuZGFLZS23,
 author = {Bo Liu and others},
 booktitle = {Proc. of NeurIPS},
 title = {{L}{I}{B}{E}{R}{O}: {B}enchmarking {K}nowledge {T}ransfer for {L}ifelong {R}obot {L}earning},
 year = {2023}
}

@article{DBLP:journals/ral/MeesHRB22,
 author = {Oier Mees and Luk{\'{a}}s Hermann and Erick Rosete{-}Beas and Wolfram Burgard},
 journal = {{IEEE} Robotics Autom. Lett.},
 pages = {7327--7334},
 title = {{C}{A}{L}{V}{I}{N}: {A} {B}enchmark for {L}anguage-{C}onditioned {P}olicy {L}earning for
{L}ong-{H}orizon {R}obot {M}anipulation {T}asks},
 year = {2022}
}

@inproceedings{DBLP:conf/corl/WalkeBZVZHHMKDL23,
 author = {Homer Rich Walke and others},
 booktitle = {Proc. of CoRL},
 pages = {1723--1736},
 title = {{B}ridge{D}ata {V}2: {A} {D}ataset for {R}obot {L}earning at {S}cale},
 year = {2023}
}

@inproceedings{DBLP:conf/rss/KhazatskyP0BDKN24,
 author = {Alexander Khazatsky and others},
 booktitle = {Proc. of RSS},
 title = {{D}{R}{O}{I}{D}: {A} {L}arge-{S}cale {I}n-{T}he-{W}ild {R}obot {M}anipulation {D}ataset},
 year = {2024}
}

@article{DBLP:journals/corr/abs-1806-10293,
 author = {Dmitry Kalashnikov and others},
 journal = {CoRR},
 title = {{Q}{T}-{O}pt: {S}calable {D}eep {R}einforcement {L}earning for {V}ision-{B}ased {R}obotic
{M}anipulation},
 year = {2018}
}

@article{DBLP:journals/ral/Corey23,
 author = {Lynch, Corey and Wahid, Ayzaan and Tompson, Jonathan and Ding, Tianli and Betker, James and Baruch, Robert and Armstrong, Travis and Florence, Pete and others},
 journal = {{IEEE} Robotics Autom. Lett.},
 pages = {1-8},
 title = {{I}nteractive {L}anguage: {T}alking to {R}obots in {R}eal {T}ime},
 year = {2023}
}

@inproceedings{DBLP:conf/corl/JangIKKELLF21,
 author = {Eric Jang and others},
 booktitle = {Proc. of CoRL},
 pages = {991--1002},
 title = {{B}{C}-{Z}: {Z}ero-{S}hot {T}ask {G}eneralization with {R}obotic {I}mitation {L}earning},
 year = {2021}
}

@inproceedings{DBLP:conf/rss/HeoLLL23,
 author = {Minho Heo and others},
 booktitle = {Proc. of RSS},
 title = {{F}urniture{B}ench: {R}eproducible {R}eal-{W}orld {B}enchmark for {L}ong-{H}orizon
{C}omplex {M}anipulation},
 year = {2023}
}

@article{DBLP:journals/ijrr/LuoXLTLWAL25,
 author = {Jianlan Luo and others},
 journal = {Int. J. Robotics Res.},
 pages = {592--606},
 title = {{F}{M}{B}: {A} functional manipulation benchmark for generalizable robotic
learning},
 year = {2025}
}

@inproceedings{DBLP:conf/corl/ShahMZ23,
 author = {Rutav Shah and Roberto Mart{\'{\i}}n{-}Mart{\'{\i}}n and Yuke Zhu},
 booktitle = {Proc. of CoRL},
 pages = {2663--2682},
 title = {{M}{U}{T}{E}{X}: {L}earning {U}nified {P}olicies from {M}ultimodal {T}ask {S}pecifications},
 year = {2023}
}

@inproceedings{DBLP:conf/corl/BelkhaleCS23,
 author = {Suneel Belkhale and others},
 booktitle = {Proc. of CoRL},
 pages = {2113--2133},
 title = {{H}{Y}{D}{R}{A}: {H}ybrid {R}obot {A}ctions for {I}mitation {L}earning},
 year = {2023}
}

@inproceedings{DBLP:conf/corl/NasirianyGMZ22,
 author = {Soroush Nasiriany and others},
 booktitle = {Proc. of CoRL},
 pages = {2181--2204},
 title = {{L}earning and {R}etrieval from {P}rior {D}ata for {S}kill-based {I}mitation {L}earning},
 year = {2022}
}

@inproceedings{DBLP:conf/icra/ZhouDSRPHJYAPFG23,
 author = {Gaoyue Zhou and others},
 booktitle = {Proc. of ICRA},
 pages = {9197--9203},
 title = {{T}rain {O}ffline, {T}est {O}nline: {A} {R}eal {R}obot {L}earning {B}enchmark},
 year = {2023}
}

@inproceedings{DBLP:conf/rss/LiuNZBZ23,
 author = {Huihan Liu and others},
 booktitle = {Proc. of RSS},
 title = {{R}obot {L}earning on the {J}ob: {H}uman-in-the-{L}oop {A}utonomy and {L}earning
{D}uring {D}eployment},
 year = {2023}
}

@inproceedings{DBLP:conf/rss/MendoncaBP23,
 author = {Russell Mendonca and others},
 booktitle = {Proc. of RSS},
 title = {{S}tructured {W}orld {M}odels from {H}uman {V}ideos},
 year = {2023}
}

@inproceedings{DBLP:conf/corl/Rosete-BeasMKBB22,
 author = {Erick Rosete{-}Beas and others},
 booktitle = {Proc. of CoRL},
 pages = {1838--1849},
 title = {{L}atent {P}lans for {T}ask-{A}gnostic {O}ffline {R}einforcement {L}earning},
 year = {2022}
}

@inproceedings{DBLP:conf/icra/MeesBB23,
 author = {Oier Mees and others},
 booktitle = {Proc. of ICRA},
 pages = {11576--11582},
 title = {{G}rounding {L}anguage with {V}isual {A}ffordances over {U}nstructured {D}ata},
 year = {2023}
}

@inproceedings{DBLP:conf/corl/MandlekarZGBSTG18,
 author = {Ajay Mandlekar and others},
 booktitle = {Proc. of CoRL},
 pages = {879--893},
 title = {{R}{O}{B}{O}{T}{U}{R}{K}: {A} {C}rowdsourcing {P}latform for {R}obotic {S}kill {L}earning
through {I}mitation},
 year = {2018}
}

@inproceedings{DBLP:conf/corl/SaxenaSK23,
 author = {Saumya Saxena and others},
 booktitle = {Proc. of CoRL},
 pages = {2210--2228},
 title = {{M}ulti-{R}esolution {S}ensing for {R}eal-{T}ime {C}ontrol with {V}ision-{L}anguage
{M}odels},
 year = {2023}
}

@misc{BerkeleyUR5Website,
 author = {Lawrence Yunliang Chen and others},
 title = {{B}erkeley {U}{R}5 {D}emonstration {D}ataset}
}

@software{dass2023jacoplay,
  author = {Dass, Shivin and Yapeter, Jullian and Zhang, Jesse and Zhang, Jiahui and Pertsch, Karl and Nikolaidis, Stefanos and Lim, Joseph J. and others},
  title = {{C}{L}{V}{R} {J}aco {P}lay {D}ataset},
  year = {2023}
}

@article{fanuc_manipulation2023,
 author = {Zhu, Xinghao and Tian, Ran and Xu, Chenfeng and Ding, Mingyu and Zhan, Wei and Tomizuka, Masayoshi and others},
 title = {{F}anuc {M}anipulation: {A} {D}ataset for {L}earning-based {M}anipulation with {F}{A}{N}{U}{C} {M}ate 200i{D} {R}obot},
 year = {2023}
}

@article{DBLP:journals/corr/abs-2210-11339,
 author = {Yifeng Zhu and others},
 journal = {CoRR},
 title = {{V}{I}{O}{L}{A}: {I}mitation {L}earning for {V}ision-{B}ased {M}anipulation with {O}bject
{P}roposal {P}riors},
 year = {2022}
}

@article{DBLP:journals/corr/abs-2307-08927,
 author = {Jianlan Luo and others},
 journal = {CoRR},
 title = {{M}ulti-{S}tage {C}able {R}outing through {H}ierarchical {I}mitation {L}earning},
 year = {2023}
}

@inproceedings{DBLP:conf/iclr/CuiWSP23,
 author = {Zichen Jeff Cui and others},
 booktitle = {Proc. of ICLR},
 title = {{F}rom {P}lay to {P}olicy: {C}onditional {B}ehavior {G}eneration from {U}ncurated
{R}obot {D}ata},
 year = {2023}
}

@article{DBLP:journals/ral/ZhuSZ22,
 author = {Yifeng Zhu and others},
 journal = {{IEEE} Robotics Autom. Lett.},
 pages = {4126--4133},
 title = {{B}ottom-{U}p {S}kill {D}iscovery {F}rom {U}nsegmented {D}emonstrations for {L}ong-{H}orizon
{R}obot {M}anipulation},
 year = {2022}
}

@inproceedings{DBLP:conf/icra/QuereHIBLSV20,
 author = {Gabriel Quere and others},
 booktitle = {Proc. of ICRA},
 pages = {1956--1962},
 title = {{S}hared {C}ontrol {T}emplates for {A}ssistive {R}obotics},
 year = {2020}
}

@article{ucsd_kitchens,
 author = {Ge Yan and others},
 title = {ucsd kitchens {D}ataset},
 year = {2023}
}

@inproceedings{DBLP:conf/cvpr/ZhaoLKFZWLMHFHL25,
 author = {Qingqing Zhao and others},
 booktitle = {Proc. of CVPR},
 pages = {1702--1713},
 title = {{C}o{T}-{V}{L}{A}: {V}isual {C}hain-of-{T}hought {R}easoning for {V}ision-{L}anguage-{A}ction
{M}odels},
 year = {2025}
}

@inproceedings{DBLP:conf/iclr/TianYZWL0P25,
 author = {Yang Tian and others},
 booktitle = {Proc. of ICLR},
 title = {{P}redictive {I}nverse {D}ynamics {M}odels are {S}calable {L}earners for {R}obotic
{M}anipulation},
 year = {2025}
}

@inproceedings{DBLP:conf/rss/GhoshWPBMDHK0LT24,
 author = {Dibya Ghosh and others},
 booktitle = {Robotics: Science and Systems XX, Delft, The Netherlands, July 15-19, 2024},
 title = {{O}cto: {A}n {O}pen-{S}ource {G}eneralist {R}obot {P}olicy},
 year = {2024}
}

@inproceedings{DBLP:journals/corr/abs-2506-19850,
 author = {Yuqi Wang and others},
 booktitle = {Proc. of ICLR},
 title = {{U}nified {V}ision-{L}anguage-{A}ction {M}odel},
 year = {2026}
}

@article{DBLP:journals/corr/abs-2506-21539,
 author = {Jun Cen and others},
 journal = {CoRR},
 title = {{W}orld{V}{L}{A}: {T}owards {A}utoregressive {A}ction {W}orld {M}odel},
 year = {2025}
}

@article{DBLP:journals/corr/abs-2508.18269,
 author = {Zhide Zhong and others},
 journal = {CoRR},
 title = {{F}low{V}{L}{A}: {T}hinking in {M}otion with a {V}isual {C}hain of {T}hought},
 year = {2025}
}

@inproceedings{DBLP:journals/rss/abs-2501-15830,
 author = {Delin Qu and others},
 booktitle = {Proc. of RSS},
 title = {{S}patial{V}{L}{A}: {E}xploring {S}patial {R}epresentations for {V}isual-{L}anguage-{A}ction
{M}odel},
 year = {2025}
}

@article{DBLP:journals/corr/abs-2412-14058,
 author = {Xinghang Li and others},
 journal = {CoRR},
 title = {{T}owards {G}eneralist {R}obot {P}olicies: {W}hat {M}atters in {B}uilding {V}ision-{L}anguage-{A}ction
{M}odels},
 year = {2024}
}

@inproceedings{DBLP:conf/iclr/ZhaoD0GB0W25,
 author = {Wei Zhao and others},
 booktitle = {Proc. of ICLR},
 title = {{V}{L}{A}{S}: {V}ision-{L}anguage-{A}ction {M}odel with {S}peech {I}nstructions for
{C}ustomized {R}obot {M}anipulation},
 year = {2025}
}

@article{DBLP:journals/corr/abs-2410-08001,
 author = {Qingwen Bu and others},
 journal = {CoRR},
 title = {{T}owards {S}ynergistic, {G}eneralized, and {E}fficient {D}ual-{S}ystem for {R}obotic
{M}anipulation},
 year = {2024}
}

@article{DBLP:journals/corr/abs-2410-24164,
 author = {Kevin Black and others},
 journal = {CoRR},
 title = {\(\pi\)\(_{\mbox{0}}\): {A} {V}ision-{L}anguage-{A}ction {F}low {M}odel for {G}eneral {R}obot {C}ontrol},
 year = {2024}
}

@article{DBLP:journals/corr/abs-2502-19645,
 author = {Moo Jin Kim and others},
 journal = {CoRR},
 title = {{F}ine-{T}uning {V}ision-{L}anguage-{A}ction {M}odels: {O}ptimizing {S}peed and {S}uccess},
 year = {2025}
}

@article{DBLP:journals/corr/abs-2505-03912,
 author = {Can Cui and others},
 journal = {CoRR},
 title = {{O}pen{H}elix: {A} {S}hort {S}urvey, {E}mpirical {A}nalysis, and {O}pen-{S}ource {D}ual-{S}ystem
{V}{L}{A} {M}odel for {R}obotic {M}anipulation},
 year = {2025}
}

@inproceedings{DBLP:journals/corr/abs-2507-04447,
 author = {Wenyao Zhang and others},
 booktitle = {Proc. of NeurIPS},
 title = {{D}ream{V}{L}{A}: {A} {V}ision-{L}anguage-{A}ction {M}odel {D}reamed with {C}omprehensive
{W}orld {K}nowledge},
 year = {2025}
}

@article{DBLP:journals/corr/abs-2508-10333,
 author = {Wenxuan Song and others},
 journal = {CoRR},
 title = {{R}econ{V}{L}{A}: {R}econstructive {V}ision-{L}anguage-{A}ction {M}odel as {E}ffective
{R}obot {P}erceiver},
 year = {2025}
}

@inproceedings{DBLP:journals/corr/abs-2509-09372,
 author = {Yihao Wang and others},
 booktitle = {Proc. of AAAI},
 title = {{V}{L}{A}-{A}dapter: {A}n {E}ffective {P}aradigm for {T}iny-{S}cale {V}ision-{L}anguage-{A}ction
{M}odel},
 year = {2026}
}

@article{DBLP:journals/corr/abs-2410-15959,
 author = {Zhi Hou and others},
 journal = {CoRR},
 title = {{D}iffusion {T}ransformer {P}olicy},
 year = {2024}
}

@article{DBLP:journals/corr/abs-2501-09747,
 author = {Karl Pertsch and others},
 journal = {CoRR},
 title = {{F}{A}{S}{T}: {E}fficient {A}ction {T}okenization for {V}ision-{L}anguage-{A}ction {M}odels},
 year = {2025}
}

@article{DBLP:journals/corr/abs-2503-14734,
 author = {Johan Bjorck and Fernando Casta{\~{n}}eda and Nikita Cherniadev and Xingye Da and Runyu Ding and Linxi Fan and Yu Fang and Dieter Fox and Fengyuan Hu and Spencer Huang and Joel Jang and Zhenyu Jiang and Jan Kautz and Kaushil Kundalia and Lawrence Lao and Zhiqi Li and Zongyu Lin and Kevin Lin and Guilin Liu and Edith LLontop and Loic Magne and Ajay Mandlekar and Avnish Narayan and Soroush Nasiriany and Scott Reed and You Liang Tan and Guanzhi Wang and Zu Wang and Jing Wang and Qi Wang and Jiannan Xiang and Yuqi Xie and Yinzhen Xu and Zhenjia Xu and Seonghyeon Ye and Zhiding Yu and Ao Zhang and Hao Zhang and Yizhou Zhao and Ruijie Zheng and Yuke Zhu},
 journal = {CoRR},
 title = {{G}{R}00{T} {N}1: {A}n {O}pen {F}oundation {M}odel for {G}eneralist {H}umanoid {R}obots},
 year = {2025}
}

@article{DBLP:journals/corr/abs-2508-20072,
 author = {Zhixuan Liang and others},
 journal = {CoRR},
 title = {{D}iscrete {D}iffusion {V}{L}{A}: {B}ringing {D}iscrete {D}iffusion to {A}ction {D}ecoding
in {V}ision-{L}anguage-{A}ction {P}olicies},
 year = {2025}
}

@article{DBLP:journals/corr/abs-2509-06951,
 author = {Qi Lv and others},
 journal = {CoRR},
 title = {{F}1: {A} {V}ision-{L}anguage-{A}ction {M}odel {B}ridging {U}nderstanding and
{G}eneration to {A}ctions},
 year = {2025}
}

@article{DBLP:journals/corr/abs-2504-16054,
 author = {Physical Intelligence and others},
 journal = {CoRR},
 title = {\(\pi\)\(_{\mbox{0.5}}\): a {V}ision-{L}anguage-{A}ction {M}odel with {O}pen-{W}orld {G}eneralization},
 year = {2025}
}

@inproceedings{DBLP:journals/corr/abs-2511-01718,
 author = {Jiayi Chen and others},
 booktitle = {Proc. of ICLR},
 title = {{U}nified {D}iffusion {V}{L}{A}: {V}ision-{L}anguage-{A}ction {M}odel via {J}oint {D}iscrete {D}enoising {D}iffusion {P}rocess},
 year = {2026}
}

@inproceedings{DBLP:conf/iclr/LoshchilovH19,
 author = {Ilya Loshchilov and Frank Hutter},
 booktitle = {Proc. of ICLR},
 title = {{D}ecoupled {W}eight {D}ecay {R}egularization},
 year = {2019}
}

@inproceedings{DBLP:conf/iclr/YuLGVSMCGGHG0ER24,
 author = {Lijun Yu and Jos{\'{e}} Lezama and Nitesh Bharadwaj Gundavarapu and Luca Versari and Kihyuk Sohn and David Minnen and Yong Cheng and Agrim Gupta and Xiuye Gu and Alexander G. Hauptmann and Boqing Gong and Ming{-}Hsuan Yang and Irfan Essa and David A. Ross and Lu Jiang},
 booktitle = {Proc. of ICLR},
 title = {{L}anguage {M}odel {B}eats {D}iffusion - {T}okenizer is key to visual generation},
 year = {2024}
}

@inproceedings{yang2025mmada,
 author = {Ling Yang and others},
 booktitle = {Proc. of NeurIPS},
 title = {{M}{M}a{D}{A}: {M}ultimodal {L}arge {D}iffusion {L}anguage {M}odels},
 year = {2025}
}

@inproceedings{nie2025large,
 author = {Shen Nie and others},
 booktitle = {Proc. of NeurIPS},
 title = {{L}arge {L}anguage {D}iffusion {M}odels},
 year = {2025}
}

@inproceedings{DBLP:conf/iclr/XieMBZWLGCYS25,
 author = {Jinheng Xie and others},
 booktitle = {Proc. of ICLR},
 title = {{S}how-o: {O}ne {S}ingle {T}ransformer to {U}nify {M}ultimodal {U}nderstanding and
{G}eneration},
 year = {2025}
}

@article{cen2025rynnvla002unifiedvisionlanguageactionworld,
 author = {Jun Cen and others},
 journal = {CoRR},
 title = {{R}ynn{V}{L}{A}-002: {A} {U}nified {V}ision-{L}anguage-{A}ction and {W}orld {M}odel},
 year = {2025}
}

@article{DBLP:journals/corr/abs-2509-06932,
 author = {Yuqing Wen and others},
 journal = {CoRR},
 title = {{L}{L}a{D}{A}-{V}{L}{A}: {V}ision {L}anguage {D}iffusion {A}ction {M}odels},
 year = {2025}
}

@inproceedings{DBLP:conf/nips/YueWKHWSF024,
 author = {Yang Yue and others},
 booktitle = {Proc. of NeurIPS},
 title = {{D}ee{R}-{V}{L}{A}: {D}ynamic {I}nference of {M}ultimodal {L}arge {L}anguage {M}odels for
{E}fficient {R}obot {E}xecution},
 year = {2024}
}

@inproceedings{DBLP:conf/kdd/RasleyRRH20,
 author = {Jeff Rasley and others},
 booktitle = {Proc. of KDD},
 pages = {3505--3506},
 title = {{D}eep{S}peed: {S}ystem {O}ptimizations {E}nable {T}raining {D}eep {L}earning {M}odels
with {O}ver 100 {B}illion {P}arameters},
 year = {2020}
}

@inproceedings{DBLP:conf/sc/RajbhandariRRH20,
 author = {Samyam Rajbhandari and others},
 booktitle = {Proceedings of the International Conference for High Performance Computing, Networking, Storage and Analysis, SC 2020, Virtual Event / Atlanta, Georgia, USA, November 9-19, 2020},
 pages = {20},
 title = {{Z}e{R}{O}: memory optimizations toward training trillion parameter models},
 year = {2020}
}

@article{DBLP:journals/corr/abs-2506-06295,
 author = {Zhiyuan Liu and others},
 journal = {CoRR},
 title = {d{L}{L}{M}-{C}ache: {A}ccelerating {D}iffusion {L}arge {L}anguage {M}odels with {A}daptive
{C}aching},
 year = {2025}
}

@inproceedings{DBLP:conf/cvpr/ChangZJLF22,
 author = {Huiwen Chang and others},
 booktitle = {Proc. of CVPR},
 pages = {11305--11315},
 title = {{M}ask{G}{I}{T}: {M}asked {G}enerative {I}mage {T}ransformer},
 year = {2022}
}

@inproceedings{DBLP:conf/iclr/BerglundTKBSKE24,
 author = {Lukas Berglund and others},
 booktitle = {Proc. of ICLR},
 title = {{T}he {R}eversal {C}urse: {L}{L}{M}s trained on "{A} is {B}" fail to learn "{B} is {A}"},
 year = {2024}
}

@article{DBLP:journals/corr/abs-2505-16839,
 author = {Shufan Li and others},
 journal = {CoRR},
 title = {{L}a{V}i{D}a: {A} {L}arge {D}iffusion {L}anguage {M}odel for {M}ultimodal {U}nderstanding},
 year = {2025}
}

@article{DBLP:journals/corr/abs-2505-16933,
 author = {Zebin You and others},
 journal = {CoRR},
 title = {{L}{L}a{D}{A}-{V}: {L}arge {L}anguage {D}iffusion {M}odels with {V}isual {I}nstruction {T}uning},
 year = {2025}
}

@article{DBLP:journals/corr/abs-2505-16990,
 author = {Runpeng Yu and others},
 journal = {CoRR},
 title = {{D}imple: {D}iscrete {D}iffusion {M}ultimodal {L}arge {L}anguage {M}odel with {P}arallel
{D}ecoding},
 year = {2025}
}

@article{DBLP:journals/corr/abs-2505-19223,
 author = {Fengqi Zhu and others},
 journal = {CoRR},
 title = {{L}{L}a{D}{A} 1.5: {V}ariance-{R}educed {P}reference {O}ptimization for {L}arge {L}anguage
{D}iffusion {M}odels},
 year = {2025}
}

@article{DBLP:journals/corr/abs-2508-15487,
 author = {Jiacheng Ye and others},
 journal = {CoRR},
 title = {{D}ream 7{B}: {D}iffusion {L}arge {L}anguage {M}odels},
 year = {2025}
}

@article{DBLP:journals/corr/abs-2510-04146,
 author = {Minseo Kim and others},
 journal = {CoRR},
 title = {{B}eyond {N}ext-{T}oken {P}rediction: {A} {P}erformance {C}haracterization of
{D}iffusion versus {A}utoregressive {L}anguage {M}odels},
 year = {2025}
}

@article{DBLP:journals/corr/abs-2509-24389,
 author = {Fengqi Zhu and others},
 journal = {CoRR},
 title = {{L}{L}a{D}{A}-{M}o{E}: {A} {S}parse {M}o{E} {D}iffusion {L}anguage {M}odel},
 year = {2025}
}

@article{DBLP:journals/corr/abs-2510-06303,
 author = {Shuang Cheng and others},
 journal = {CoRR},
 title = {{S}{D}{A}{R}: {A} {S}ynergistic {D}iffusion-{A}uto{R}egression {P}aradigm for {S}calable
{S}equence {G}eneration},
 year = {2025}
}

@inproceedings{Radford2018ImprovingLU,
 author = {Alec Radford and Karthik Narasimhan},
 title = {{I}mproving {L}anguage {U}nderstanding by {G}enerative {P}re-{T}raining},
 year = {2018}
}

@inproceedings{Radford2019LanguageMA,
 author = {Alec Radford and others},
 title = {{L}anguage {M}odels are {U}nsupervised {M}ultitask {L}earners},
 year = {2019}
}

@inproceedings{conf/nips/BrownMRSKDNSSAA20,
 author = {Tom B. Brown and others},
 booktitle = {Proc. of NeurIPS},
 title = {{L}anguage {M}odels are {F}ew-{S}hot {L}earners},
 year = {2020}
}

@article{journals/corr/abs-2302-13971,
 author = {Hugo Touvron and others},
 journal = {CoRR},
 title = {{L}{L}a{M}{A}: {O}pen and {E}fficient {F}oundation {L}anguage {M}odels},
 year = {2023}
}

@article{DBLP:journals/corr/abs-2309-16609,
 author = {Jinze Bai and others},
 journal = {CoRR},
 title = {{Q}wen {T}echnical {R}eport},
 year = {2023}
}

@article{DBLP:journals/corr/abs-2501-12948,
 author = {Daya Guo and others},
 journal = {Nature},
 pages = {633--638},
 title = {{D}eep{S}eek-{R}1 incentivizes reasoning in {L}{L}{M}s through reinforcement learning},
 year = {2025}
}

@inproceedings{DBLP:conf/iccv/ZhangRA23,
 author = {Lvmin Zhang and others},
 booktitle = {Proc. of ICCV},
 pages = {3813--3824},
 title = {{A}dding {C}onditional {C}ontrol to {T}ext-to-{I}mage {D}iffusion {M}odels},
 year = {2023}
}

@inproceedings{DBLP:conf/iclr/BenitaEK24,
 author = {Roi Benita and others},
 booktitle = {Proc. of ICLR},
 title = {{D}iff{A}{R}: {D}enoising {D}iffusion {A}utoregressive {M}odel for {R}aw {S}peech {W}aveform
{G}eneration},
 year = {2024}
}

@inproceedings{DBLP:conf/nips/DhariwalN21,
 author = {Prafulla Dhariwal and Alexander Quinn Nichol},
 booktitle = {Proc. of NeurIPS},
 pages = {8780--8794},
 title = {{D}iffusion {M}odels {B}eat {G}{A}{N}s on {I}mage {S}ynthesis},
 year = {2021}
}

@inproceedings{DBLP:conf/nips/HoJA20,
 author = {Jonathan Ho and others},
 booktitle = {Proc. of NeurIPS},
 title = {{D}enoising {D}iffusion {P}robabilistic {M}odels},
 year = {2020}
}

@inproceedings{DBLP:conf/nips/LiHRMM23,
 author = {Yinghao Aaron Li and others},
 booktitle = {Proc. of NeurIPS},
 title = {{S}tyle{T}{T}{S} 2: {T}owards {H}uman-{L}evel {T}ext-to-{S}peech through {S}tyle {D}iffusion
and {A}dversarial {T}raining with {L}arge {S}peech {L}anguage {M}odels},
 year = {2023}
}

@inproceedings{DBLP:conf/corl/ZitkovichYXXXXW23,
 author = {Brianna Zitkovich and others},
 booktitle = {Proc. of CoRL},
 pages = {2165--2183},
 title = {{R}{T}-2: {V}ision-{L}anguage-{A}ction {M}odels {T}ransfer {W}eb {K}nowledge to {R}obotic
{C}ontrol},
 year = {2023}
}

@inproceedings{DBLP:conf/icml/BruceDEPS0LMSAA24,
 author = {Jake Bruce and others},
 booktitle = {Proc. of ICML},
 title = {{G}enie: {G}enerative {I}nteractive {E}nvironments},
 year = {2024}
}

@article{DBLP:journals/corr/abs-2506-09985,
 author = {Mido Assran and others},
 journal = {CoRR},
 title = {{V}-{J}{E}{P}{A} 2: {S}elf-{S}upervised {V}ideo {M}odels {E}nable {U}nderstanding, {P}rediction
and {P}lanning},
 year = {2025}
}

@inproceedings{DBLP:conf/naacl/DevlinCLT19,
 author = {Jacob Devlin and others},
 booktitle = {Proc. of NAACL},
 pages = {4171--4186},
 title = {{B}{E}{R}{T}: {P}re-training of {D}eep {B}idirectional {T}ransformers for {L}anguage
{U}nderstanding},
 year = {2019}
}

@article{DBLP:journals/corr/abs-2510-09667,
 author = {Huaihai Lyu and others},
 journal = {CoRR},
 title = {{O}mni{S}{A}{T}: {C}ompact {A}ction {T}oken, {F}aster {A}uto {R}egression},
 year = {2025}
}

@inproceedings{Wang_2025_ICCV,
 author = {Wang, Yating and Zhu, Haoyi and Liu, Mingyu and Yang, Jiange and Fang, Hao-Shu and He, Tong and others},
 booktitle = {Proc. of ICCV},
 pages = {11089-11099},
 title = {{V}{Q}-{V}{L}{A}: {I}mproving {V}ision-{L}anguage-{A}ction {M}odels via {S}caling {V}ector-{Q}uantized {A}ction {T}okenizers},
 year = {2025}
}

@inproceedings{DBLP:conf/iros/SongCDZZZGLWWML25,
 author = {Wenxuan Song and others},
 booktitle = {Proc. of IROS},
 pages = {13162--13169},
 title = {{P}{D}-{V}{L}{A}: {A}ccelerating {V}ision-{L}anguage-{A}ction {M}odel {I}ntegrated with
{A}ction {C}hunking via {P}arallel {D}ecoding},
 year = {2025}
}

@article{DBLP:journals/corr/abs-2506-13725,
 author = {Wenxuan Song and others},
 journal = {CoRR},
 title = {{C}{E}{E}{D}-{V}{L}{A}: {C}onsistency {V}ision-{L}anguage-{A}ction {M}odel with {E}arly-{E}xit
{D}ecoding},
 year = {2025}
}

@inproceedings{DBLP:journals/corr/abs-2512-00975,
 author = {Haotian Liang and others},
 booktitle = {Proc. of CVPR},
 title = {{M}{M}-{A}{C}{T}: {L}earn from {M}ultimodal {P}arallel {G}eneration to {A}ct},
 year = {2026}
}

@inproceedings{DBLP:journals/corr/abs-2512-09928,
 author = {Minghui Lin and others},
 booktitle = {Proc. of CVPR},
 title = {{H}i{F}-{V}{L}{A}: {H}indsight, {I}nsight and {F}oresight through {M}otion {R}epresentation
for {V}ision-{L}anguage-{A}ction {M}odels},
 year = {2026}
}


\end{document}